\documentclass[sigconf]{acmart}
\AtBeginDocument{%
  }

\setcopyright{rightsretained}
\copyrightyear{2025}
\acmYear{2025}
\acmDOI{XXXXXXX.XXXXXXX}
\acmConference[KDD'25]{August 03-07,2025}{Toronto}{Canada}

\usepackage{rotating}
\usepackage{multirow}
\usepackage{amsthm}
\usepackage{color}
\usepackage{pifont}
\usepackage{graphicx}
\usepackage{subcaption}

\begin{document}

\title{Understanding Why Large Language Models Can Be Ineffective in Time Series Analysis: The Impact of Modality Alignment}

\author{Liangwei Nathan Zheng}
\email{liangwei.zheng@adelaide.edu.au}
\affiliation{%
  \institution{The University of Adelaide}
  \city{Adelaide}
  \state{South Australia}
  \country{Australia}
}

\author{Chang Dong}
\email{chang.dong@adelaide.edu.au}
\affiliation{%
  \institution{The University of Adelaide}
  \city{Adelaide}
  \state{South Australia}
  \country{Australia}
}

\author{Wei Emma Zhang}
\email{wei.e.zhang@adelaide.edu.au}
\affiliation{%
  \institution{The University of Adelaide}
  \city{Adelaide}
  \state{South Australia}
  \country{Australia}
}

\author{Lin Yue}
\email{lin.yue@adelaide.edu.au}
\affiliation{%
  \institution{The University of Adelaide}
  \city{Adelaide}
  \state{South Australia}
  \country{Australia}
}

\author{Miao Xu}
\email{miao.xu@uq.edu.au}
\affiliation{%
  \institution{The University of Queensland}
  \city{Brisbane}
  \state{Queensland}
  \country{Australia}
}

\author{Olaf Maennel}
\email{olaf.maennel@adelaide.edu.au}
\affiliation{%
  \institution{The University of Adelaide}
  \city{Adelaide}
  \state{South Australia}
  \country{Australia}
}

\author{Weitong Chen}
\email{weitong.chen@adelaide.edu.au}
\authornote{Corresponding Author}
\affiliation{%
  \institution{The University of Adelaide}
  \city{Adelaide}
  \state{South Australia}
  \country{Australia}
}

\renewcommand{\shortauthors}{Liangwei et al.}


\begin{abstract}
    Large Language Models (LLMs) have demonstrated impressive performance in time series analysis and seems to understand the time temporal relationship well than traditional transformer-based approaches. However, since LLMs are not designed for time series tasks, simpler models—like linear regressions can often achieve comparable performance with far less complexity. In this study, we perform extensive experiments to assess the effectiveness of applying LLMs to key time series tasks, including forecasting, classification, imputation, and anomaly detection. We compare the performance of LLMs against simpler baseline models, such as single-layer linear models and randomly initialized LLMs. Our results reveal that LLMs offer minimal advantages for these core time series tasks and may even distort the temporal structure of the data. In contrast, simpler models consistently outperform LLMs while requiring far fewer parameters. Furthermore, we analyze existing reprogramming techniques and show, through data manifold analysis, that these methods fail to effectively align time series data with language and display "pseudo-alignment" behavior in embedding space. Our findings suggest that the performance of LLM-based methods in time series tasks arises from the intrinsic characteristics and structure of time series data, rather than any meaningful alignment with the language model architecture. We release the code for experiments here: \url{https://github.com/IcurasLW/Official-Repository_Understanding_LLM_for_Time_Series_Analysis.git}
\end{abstract}




\maketitle

\section{Introduction}
The capability of Large Language Models (LLMs) spans the application to content recommendations of e-commerce, online finance analysis, and multivariate sensory data analysis\cite{wu2024surveylargelanguagemodels,zhao2024revolutionizing,lim2021time}. In particular, machine learning for multivariate time series data contains many tasks such as ECG data classification\cite{shen2022leads,dong2023swap}, ICU monitoring\cite{zheng2024irregularity,zheng2025free} and online traffic flow forecasting \cite{lim2021time,chen2018eeg}. Time series analysis method spans from statistical methods (e.g. ARIMA) to deep learning-based frameworks (e.g. Informer\cite{zhou2021informer}). Due to the great success of LLMs for natural language processing. LLMs is known to have large knowledge of the understanding of multimodal data and strong reasoning capacity \cite{alayrac2022flamingovisuallanguagemodel,yin2024surveymultimodallargelanguage}. Language is one of the common sequential data structures similar to time series, where the implication of each token temporally depends on the neighbour tokens. This intriguing intuition raises surge explorations on reprogram LLM to time series data. 

Recent studies\cite{zhou2023one,jin2023time,liu2024timecma,gruver2024large,liu2024calfaligningllmstime,pan2024textbf} explore a potential utilization of the Large Language Model for Time Series (LLM4TS) by reprogramming the LLM to adapt the underlying data manifold of time series. These LLM4TS methods demonstrate that LLMs can achieve and even outperform the time series expert network in general time series tasks such as forecasting, classification, anomaly detection and imputation. Some researchers \cite{zhou2023one,liu2024calfaligningllmstime,pan2024textbf} believe that time series data and language share a similar training scheme behind the scenes: predict the tokens for the next sequence length based on historical measurements. Previous studies \cite{zhao2023tuning,lu2022frozen} claim that the reprogramming techniques, such as fine-tuning LayerNorm and multihead attention alignment to text prototypes, can capture the underlying data manifold of time series from a language perspective. However, we argue that the impressive results come from the sparsity and internal structure of time series instead of language knowledge. In addition, many linear-based evidence such as (DLinear\cite{zeng2023transformers}, FITS\cite{xu2023fits} and OLS\cite{toner2024analysis}) can achieve SOTA performance without the complexity of LLMs on time series forecasting task. Moreover, time series data often exhibit sparse, stronger temporal correlation than language. For instance, a word at the end of the sentence may be strongly correlated with the word at the beginning of the sentence while a time reading often correlates to its adjacent readings given the fact of decaying autocorrelation \cite{shumway2000time}. Furthermore, time series data naturally exhibit stronger temporal relationships, such as seasonal and trend patterns over certain time lags, whereas language data rarely displays such clear patterns in real-world contexts. The temporal dependencies in language are more contextual and less deterministic. Words influence each other based on meaning, syntax, and grammar, rather than strictly on past values, making the attention mechanism fit more on text data than time series \cite{zeng2023transformers}. Although an existing work \cite{tan2024languagemodelsactuallyuseful} has discussed the impacts of LLM for time series data, the paper is limited to forecasting tasks with incomplete studies for other important tasks and lacks of in-depth understanding of the reprogramming mechanism of LLMs. It still remains unclear if LLMs capture the time dependency effectively as a simple model and why LLMs can perform time series tasks with reprogramming techniques.

To validate the necessity of LLMs for foundational time series tasks, we conduct extensive experiments with four state-of-the-art LLM4TS methods focused on long-term forecasting, classification, anomaly detection, and imputation. Surprisingly, our empirical results reveal that LLMs underperform compared to simpler ablation models, such as single-layer linear backbones and randomly initialized models. In addition, we systematically deconstruct existing reprogramming techniques: LayerNorm fine-tuning, text-time alignment and LoRA fine-tuning\cite{hu2021lora} to further understand the impacts of existing reprogramming LLMs approach for time series through the modality alignment perspective. \textbf{Our analysis reveals}: (1) LLMs are less effective for foundational time series tasks and not computationally worthwhile compared to simpler methods. (2) Reprogramming LLM4TS methods display ``pseudo-alignment" behavior, transferring only the centroid of time series data without altering the data manifold, which hinders the reasoning capabilities of LLMs. (3) Existing reprogramming techniques are generally less efficient in capturing temporal dependencies compared to straightforward architectures, such as single-layer linear models. (4) The performance of LLMs for TS came from the internal structure of time series data instead of language knowledge. This paper aims to investigate the effectiveness of LLMs on general time series tasks and the mechanisms behind reprogramming LLMs, providing insights for successful LLM reprogramming strategies. To sum up, the contributions of this paper can be summarized:

\begin{itemize}
    \item We reproduce the results of SOTA LLMs-based methods for time series tasks, including forecasting, imputation, classification, and anomaly detection. Our analysis reveals that \textbf{LLMs may not essential for effective time series analysis}.
    \item Our statistical analysis reveals that while LLMs can partially capture time series dependency, they are not as effective as simpler models trained from scratch.
    \item We demonstrate that the data manifold of time series exhibits significantly more variability than language, which hinders the ability of LLMs to effectively leverage the knowledge of LLM in time series analysis.
    \item To the best of our knowledge, we are the first to comprehensively validate the feasibility of LLMs across all time series tasks from the perspective of modality alignment.
\end{itemize}

\section{Related Works}
\textbf{Reprogramming LLM for Time Series Task}. Recently, LLMs have shown their strong cross-modalities learning capacity across domains and modalities \cite{alayrac2022flamingovisuallanguagemodel,zhou2023one,melnyk2023reprogramming,jin2024timellmtimeseriesforecasting}.  One-Fits-All believes LLM perform similar mechanism between language and time series data and fine-tune the positional encoding and layernorm layer to perform general time series tasks. Similarly, S2IP\cite{pan2024textbf} decomposes time series before LLM with layernorm fine-tune to reveal clear trend and seasonality features. On the other hand, LLM4TS \cite{chang2024llm4tsaligningpretrainedllms} fine-tuning the whole GPT2 model with LoRa to perform time series task. CALF\cite{liu2024calfaligningllmstime} attempt to align the time series prediction from text prototype and raw time series by consistency loss. In addition, LLMTime \cite{gruver2024largelanguagemodelszeroshot} and Time-LLM\cite{jin2024timellmtimeseriesforecasting} transfer LLM in zero-shot setting for forecasting task. Specifically, LLMTime converts each numerical reading to a series string of numbers. Time-LLM incorporates in-context learning and aligns the time series token with text prototypes by one layer cross-attention. The optimized object of reprogramming LLM for time series is to align time series data with language distribution so as that activate the powerful reasoning ability of LLM for time series tasks.

\textbf{Light-weight Model for Time Series Data}. Apart from the heavy architecture of LLM, there are light-weight methods modelling the time series data from the perspective of linearity. DLinear\cite{zeng2023transformers} challenges Transformer-based methods with attention is difficult to capture the strictly temporal relationship between time steps and defeats many Transformer-based architectures by one layer simple linear MLP. TSMixer\cite{ekambaram2023tsmixer}, follows this idea to further improve performance by mixing inter and intra time series relationship with multiple MLPs. Similarly, FITS\cite{xu2023fits} models time series with selected frequency components by MLP, achieving SOTA performance with the least parameters. OSL\cite{toner2024analysis} even abandons the deep learning framework, but statically fits an explicit solution of the linear regression function, proving that a simple Linear function can achieve SOTA performance as a complex deep learning framework. These are solid evidence that inspires us to further investigate whether an LLM is computational-worthy for time series tasks due to the high cost of resources.

\textbf{Time Series Foundation Model}
Apart from LLM, there are some researchers have started developing time series foundation models specifically trained on time series domain. TimeGPT1\cite{garza2024timegpt1} proposed the first foundation time series model in GPT-like achitecture, trained on 100 billion data samples from various domains such as finance, economics, healthcare, weather, IoT sensor data and so on. TimeGPT1 patches the time serise in a fixed windows size. Chronos\cite{ansari2024chronos} develop a T5-like architecture as time foundation model by learning the time series reading in string format and train the model in a forecasting manner. Similarly, Lag-Llama \cite{rasul2023lag} employ a decoder only LLaMa-like architecture and trained on uni-variate time series data with lagged tokenization. MOMENT\cite{goswami2024moment} utilize Transformer-encoder only architecture and trained on not just forecasting dataset but also anomaly and classification dataset. MOMENT was trained by imputing the masked time step reading, aiming to understand the time series dependency.

\begin{table*}[ht]
    \centering
    \begin{tabular}{c|ccc|cccccc}
         \hline
         \multirow{2}{*}{Method} & \multirow{2}{*}{M. Align} & \multirow{2}{*}{Finetune} & \multirow{2}{*}{Backbone} & \multicolumn{6}{c}{\#Params of Variants} \\
         & & & & LLM & Random & Linear & Att & Trans & NoLLM \\
         \hline
         OFA      & \ding{55} & \ding{51} & 6-layers GPT2   & 82.33 M   & 82.33 M   & 1.71 M  & 4.75 M  & 7.22 M  & 1.12 M  \\
         Time-LLM & \ding{51} & \ding{55} & 32-layers LLama & 6657.86 M & 6657.86 M & 92.63 M & 94.99 M & 98.14 M & 92.04 M \\
         CALF     & \ding{51} & \ding{51} & 6-layers GPT2   & 180.10 M  & 180.10 M  & 17.46 M & 22.18 M & 28.49 M & 16.28 M \\
         S2IP     & \ding{51} & \ding{51} & 6-layers GPT2   & 172.47 M  & 172.47 M  & 91.15M  & 93.52 M & 96.66 M & 90.56 M \\
         \hline
    \end{tabular}
    \caption{Model Statistics}
    \label{tab:Model Statistics}
\end{table*}

\section{Preliminaries}
\textbf{Time series modelling}. Given a time series data $\mathcal{X} = \{X_1, \dots, X_t \}^L_{t=1}$, where $L$ is the sequence length of time series. For any time step $X^t$, it contains $\mathcal{V}$ variates $\mathcal{X}^t = \{X_t^1, \dots, X_t^i \}^V_{i=1}$. The goal is to learn a strong temporal embedding $\mathbf{E}\in \mathbb{R}^{S \times D}$ for downstream tasks, where $S$ and $P$ is embedding sequence length and embedding dimension. Specifically, the forecasting task predicts the value readings in next $N$ arbitrary time step based on the historical temporal embedding $\mathbf{E}$. Imputation task aims to reconstruct the input $\mathcal{X}$ with missing value in any arbitrary time step $m^t_i$ of the input time series. The classification task aims to classify the $\mathcal{X}$ into $C$ classes. Anomaly detection task aims to find the anomaly point at $m^t_i$ given a distinguishable threshold determined from the dataset. A good anomaly detection model is expected to reconstruct the input time series via $\mathbf{E}$ in the temporal component. Anomalies are detected when the reconstruction error exceeds a predefined threshold. 

\textbf{Model Reprogramming} is a transfer learning technique that repurposes a pre-trained model without or with the least training computational efforts. Given a source model $\mathcal{M}_S$ that we wish to transfer, and source embedding distribution $E_S \in \mathcal{R}^S$. The target is to utilize the reasoning power of $\mathcal{M}_S$ to predict over target distribution $E_T \in \mathcal{R}^T$. Unlike fine-tuning and zero-shot learning, model reprogramming aims to transfer the input data in $E_T$ into $E_S$. Adversarial reprogramming fools $\mathcal{M}_S$ by adversarial noise to make predictions for the input target data without finetuning. For example, Voice2Series \cite{yang2021voice2series} add adversarial noise, generated from an audio classification model, over IoT sensor time series data for time series classification task without finetuning $\mathcal{M}_S$. It adopts a label mapping function that assigns time series labels over audio labels. On the other hand, FPT \cite{lu2022frozen} only fine-tuned the LayerNorm layer of pre-trained transformer models (e.g. GPT2, ViT, BERT) to transfer the knowledge for cross-domain tasks such as image classification, bit memory prediction, and remote homology detection. FPT believes finetuning LayerNorm can adopt the knowledge of $\mathcal{M}_S$ to $E_T$. However, finetuning LayerNorm potentially works well when $E_T$ and $E_S$ present similar data structures or are well-aligned since there is no theoretical guarantee to align $E_T$ and $E_S$. To fully utilize knowledge of $\mathcal{M}_S$, the objective is to align the target distribution $E_T$ to source distribution $E_S$. Specifically, we show the reprogramming object is to minimize the Wasserstein-1 distance between the target data distribution (Proof in Appendix \ref{sec:proof_theorem}).

\begin{theorem}
For a pre-trained LLM $\mathcal{H}(\cdot) = f_K(\cdot)$, where $f_K(\cdot)$ is the K-Lipschitz neural network. Given target time series data $t_i$ and source language data $s_i$, the empirical goal of reprogramming is defined as the minimization of Wasserstein-1 distance of $s_i$ and $t_i^*$, where $t_i^*$ is the data representation aligned to $s_i$:

\begin{equation}
    \left| E\left[ f_K(s_i) \right] - E\left[ f_K(t_i^*) \right] \right| \leq K \mathcal{W}_1 \left(s_i, t_i^*\right)
\end{equation}
\end{theorem}

\section{Experiment}
\subsection{LLM for TS Models}
\ \ \ \ \textbf{One-Fits-All\cite{zhou2023one}} construct time series tokens by patching them with a user-defined window size and directly fed the time series token into a 6-layers GPT2 after instance norm in a patch. They finetuned only the LayerNorm layers of GPT2, aiming to adapt time series distribution into language.

\textbf{Time-LLM\cite{jin2024timellmtimeseriesforecasting}} adopt the same patching technique as OFA and align the time series to text prototypes by one layer of multihead cross-attention. The aligned tokens are then concatenated to the embedding of dataset description and time series statistics prompt. The concatenated embedding is passed to a frozen 32-layer pretrained LLaMa for forecasting task.

\textbf{CALF\cite{liu2024calfaligningllmstime}} projects the text vocabulary to principle component space and fuse the principle text component with time series tokens by multihead attention. The fused features are passed to a frozen LLM for making predictions from the text-domain. The embedded time series tokens are passed to a LoRA-finetuned LLM for making predictions from temporal prediction. The predictions from both domains are expected to align in the same feature space by consistency loss.

\textbf{S2IP\cite{pan2024textbf}} learns the temporal feature from decomposed components (seasonal, trend and residual) and patches time series in three decomposed components. The decomposed components are attended to select the top-K similar semantic text prototype. These prototypes are prefixed to TS embedding, representing the description of the components. Instead of making a prediction directly, S2IP predicts decomposed components and reverses the additive decomposition back to the original series.

\subsection{Dataset and Reproduction}
For a fair comparison, we evaluate the above LLM4TS models on multiple benchmark datasets, which are used in the experiments of One-Fits-All\cite{zhou2023one} and Time Series Library Benchmark\cite{wang2024tssurvey}. (1) Long-term forecasting: \textbf{ETT (m1, m2, h1, h2)}, \textbf{ECL}, \textbf{Weather}, \textbf{Illness}, \textbf{Traffic}. (2) Imputation: \textbf{ETT (m1, m2, h1, h2)}, \textbf{ECL} and \textbf{Weather}. (3) Anomaly Detection: \textbf{SMD}, \textbf{SWaT}, \textbf{PSM}, \textbf{SMAP} and \textbf{MSL}. (4) Classification: 10 \textbf{UEA} datasets used in One-Fits-All experiments. As Time-LLM, CALF, and S2IP only implement time series forecasting task in the original implementation. We adopt the dataset setting from OFA and the best parameter they used in forecasting task to finetune the best learning rate for convergence. All experiments are reproduced on 4 $\times$ RTX4090 24GB, 1 $\times$ A6000 48GB, and 4 $\times$ A100-SXM4-40GB. For S2IP and TimeLLM, we accelerate their data processing procedure by data parallel processing. In the original implementation, S2IP and Time-LLM generate data in the model forward function, which uses only one CPU processor to process the batch data and significantly slow down the training speed without fully utilising GPU capacity. We implement the data processing procedure of S2IP and Time-LLM in the Torch DataLoader class for parallel data processing. We give detailed reproduction notes in Appendix \ref{sec:reproduction note}.

\subsection{Model Variants}
\textbf{Random} is a model variant where the LLM parameters are re-initialized, and only the LayerNorm layer is fine-tuned. In this setup, the attention and feedforward layers encapsulate the LLM's language knowledge. This approach allows for an investigation into whether the pre-existing language knowledge is truly beneficial for time series tasks.

\textbf{Linear} is a model variant where the LLM backbone is replaced by a simple linear and LayerNorm layer. This is motivated by previous lightweight works such as DLinear\cite{zeng2023transformers} and FITS\cite{xu2023fits}. We hypothesize that the inherent sparsity of time series data makes it more conducive to learning through linear functions, offering a more efficient approach compared to complex LLM architectures.

\textbf{Att} and \textbf{Trans} are two variants that replace the LLM backbone with a layer of multi-head Attention and Transformer Encoder. These variants aim to examine if simple attention and transformer encoder layer can achieve similar results as complicated LLM.

\textbf{NoLLM} is the variant that removes LLM backbone and passes the input embedding directly to output projection, which examines LLM architecture contributes to time series modeling. This variant aims to investigate if LLM structure is necessary for time series modelling.

\begin{table*}[ht]
    \centering
    \scalebox{0.9}{
    \begin{tabular}{c|cc|cc|cc|cc|cc|cc|cc}
         \hline
         \multicolumn{15}{c}{One-Fits-ALL} \\
         \hline
         \multirow{2}{*}{Dataset} & \multicolumn{2}{c|}{LLM} &  \multicolumn{2}{c|}{Random} & \multicolumn{2}{c|}{LN} & \multicolumn{2}{c|}{Att} & \multicolumn{2}{c|}{Trans} & \multicolumn{2}{c|}{NoLLM} & \multicolumn{2}{c}{Original}\\
        & MSE & MAE & MSE & MAE & MSE & MAE & MSE & MAE & MSE & MAE & MSE & MAE & MSE & MAE \\
         \hline
         
        ETTh1   & 0.433 & \textcolor{blue}{\textbf{0.435}} & 0.508 & 0.479 & 0.432 & 0.436 & 0.486 & 0.452 & 0.504 & 0.473 & \textcolor{red}{\textbf{0.431}} & 0.437 & 0.427 & 0.426 \\
        ETTh2   & 0.362 & 0.399 & 0.433 & 0.448 & 0.361 & \textcolor{blue}{\textbf{0.397}} & 0.363 & 0.401 & 0.404 & 0.433 & \textcolor{red}{\textbf{0.356}} & 0.398 & 0.346 & 0.394 \\
        ETTm1   & \textcolor{red}{\textbf{0.360}} & 0.394 & 0.367 & 0.392 & 0.372 & 0.386 & 0.367 & 0.388 & 0.388 & 0.406 & 0.364 & \textcolor{blue}{\textbf{0.383}} & 0.352 & 0.383 \\
        ETTm2   & 0.272 & 0.332 & 0.292 & 0.346 & 0.270 & 0.326 & 0.261 & 0.322 & 0.279 & 0.332 & \textcolor{red}{\textbf{0.259}} & \textcolor{blue}{\textbf{0.317}} & 0.266 & 0.326 \\
        Illness & 2.209 & 0.981 & 1.961 & 0.920 & \textcolor{red}{\textbf{1.901}} & \textcolor{blue}{\textbf{0.891}} & 2.209 & 0.960 & 2.004 & 0.928 & 2.134 & 0.981 & 1.925 & 0.903 \\
        Weather & \textcolor{red}{\textbf{0.219}} & \textcolor{blue}{\textbf{0.265}} & 0.243 & 0.283 & 0.221 & 0.266 & 0.221 & 0.266 & 0.225 & 0.270 & 0.249 & 0.281 & 0.237 & 0.270 \\
        Traffic & 0.428 & 0.296 & 0.407 & 0.281 & 0.420 & 0.292 & 1.061 & 0.629 & \textcolor{red}{\textbf{0.401}} & \textcolor{blue}{\textbf{0.278}} & 0.422 & 0.287 & 0.414 & 0.294 \\
        ECL     & 0.167 & 0.266 & \textcolor{red}{\textbf{0.163}} & \textcolor{blue}{\textbf{0.255}} & 0.164 & 0.258 & 0.165 & 0.260 & 0.164 & 0.261 & 0.164 & 0.257 & 0.167 & 0.263 \\
        \hline
        Wins  & 2 & 2 & 1 & 1 & 1 & 2 & 0 & 0 & 1 & 1 & 3 & 2 & - & - \\
         \hline

         \multicolumn{15}{c}{Time-LLM} \\
         \hline
         \multirow{2}{*}{Dataset} & \multicolumn{2}{c|}{LLM} &  \multicolumn{2}{c|}{Random} & \multicolumn{2}{c|}{LN} & \multicolumn{2}{c|}{Att} & \multicolumn{2}{c|}{Trans} & \multicolumn{2}{c|}{NoLLM} & \multicolumn{2}{c}{Original} \\
        & MSE & MAE & MSE & MAE & MSE & MAE & MSE & MAE & MSE & MAE & MSE & MAE & MSE & MAE\\
         \hline
        ETTh1   & 0.426 & 0.439 & 0.417 & 0.437 & 0.426 & 0.441 & 0.424 & 0.440 & 0.438 & 0.449 & \textcolor{red}{\textbf{0.411}} & \textcolor{blue}{\textbf{0.431}} & 0.408 & 0.423 \\
        ETTh2   & 0.355 & 0.403 & 0.361 & 0.406 & 0.362 & 0.403 & \textcolor{red}{\textbf{0.354}} & \textcolor{blue}{\textbf{0.386}} & 0.359 & 0.393 & 0.358 & 0.395 & 0.334 & 0.383 \\
        ETTm1   & 0.361 & 0.389 & \textcolor{red}{\textbf{0.352}} & \textcolor{blue}{\textbf{0.384}} & 0.364 & 0.395 & 0.367 & 0.395 & 0.360 & 0.390 & 0.354 & 0.389 & 0.329 & 0.372 \\
        ETTm2   & 0.267 & 0.328 & \textcolor{red}{\textbf{0.266}} & \textcolor{blue}{\textbf{0.325}} & 0.267 & 0.329 & 0.274 & 0.340 & 0.274 & 0.335 & 0.271 & 0.333 & 0.251 & 0.313 \\
        Illness & 1.857 & 0.907 & 1.854 & 0.920 & 1.788 & 0.861 & \textcolor{red}{\textbf{1.719}} & \textcolor{blue}{\textbf{0.822}} & 1.775 & 0.866 & 1.870 & 0.947 & - & -\\
        Weather & 0.251 & 0.266 & 0.236 & \textcolor{blue}{\textbf{0.261}} & \textcolor{red}{\textbf{0.226}} & 0.263 & 0.236 & 0.268 & 0.240 & 0.276 & 0.250 & 0.273 & 0.225 & 0.257 \\
        Traffic & 0.424 & 0.283 & 0.425 & 0.285 & \textcolor{red}{\textbf{0.413}} & \textcolor{blue}{\textbf{0.281}} & 0.420 & 0.285 & 0.421 & 0.282 & 0.434 & 0.301 & 0.388 & 0.264 \\
        ECL     & 0.165 & 0.260 & 0.169 & 0.265 & 0.178 & 0.273 & 0.168 & 0.262 & \textcolor{red}{\textbf{0.161}} & \textcolor{blue}{\textbf{0.254}} & 0.175 & 0.270 & 0.158 & 0.252 \\
        \hline
        Wins  & 0 & 0 & 2 & 3 & 2 & 1 & 2 & 2 & 1 & 1 & 1 & 1 & - & - \\
         \hline
         
        \multicolumn{15}{c}{CALF} \\
        \hline
         \multirow{2}{*}{Dataset} & \multicolumn{2}{c|}{LLM} &  \multicolumn{2}{c|}{Random} & \multicolumn{2}{c|}{LN} & \multicolumn{2}{c|}{Att} & \multicolumn{2}{c|}{Trans} & \multicolumn{2}{c|}{NoLLM} & \multicolumn{2}{c}{Original} \\
        & MSE & MAE & MSE & MAE & MSE & MAE & MSE & MAE & MSE & MAE & MSE & MAE & MSE & MAE\\
         \hline
        ETTh1   & 0.439 & 0.434 & 0.446 & 0.436 & \textcolor{red}{\textbf{0.438}} & \textcolor{blue}{\textbf{0.429}} & 0.440 & 0.434 & 0.439 & 0.430 & 0.440 & 0.432 & 0.432 & 0.428 \\
        ETTh2   & 0.376 & 0.398 & 0.375 & 0.397 & 0.373 & \textcolor{blue}{\textbf{0.395}} & \textcolor{red}{\textbf{0.372}} & 0.395 & 0.374 & 0.395 & 0.375 & 0.397 & 0.349 & 0.382 \\
        ETTm1   & 0.393 & 0.390 & 0.396 & 0.391 & 0.383 & 0.383 & 0.384 & 0.384 & \textcolor{red}{\textbf{0.383}} & \textcolor{blue}{\textbf{0.382}} & 0.386 & 0.386 & 0.395 & 0.390 \\
        ETTm2   & 0.282 & 0.322 & 0.281 & 0.320 & 0.278 & 0.318 & 0.280 & 0.320 & \textcolor{red}{\textbf{0.278}} & \textcolor{blue}{\textbf{0.318}} & 0.280 & 0.320 & 0.281 & 0.321 \\
        Illness & 2.361 & 1.035 & 2.079 & 0.960 & \textcolor{red}{\textbf{2.001}} & 0.930 & 2.196 & 0.991 & 2.079 & \textcolor{blue}{\textbf{0.952}} & 2.154 & 0.982 & - & - \\
        Weather & \textcolor{red}{\textbf{0.246}} & \textcolor{blue}{\textbf{0.271}} & 0.255 & 0.277 & 0.261 & 0.280 & 0.258 & 0.278 & 0.257 & 0.277 & 0.258 & 0.277 & 0.250 & 0.274 \\
        Traffic & 0.459 & 0.303 & 0.459 & 0.306 & \textcolor{red}{\textbf{0.456}} & 0.299 & 0.466 & 0.304 & 0.455 & \textcolor{blue}{\textbf{0.298}} & 0.463 & 0.306 & 0.439 & 0.281 \\
        ECL     & 0.193 & 0.278 & 0.189 & 0.275 & 0.186 & 0.271 & 0.188 & 0.275 & \textcolor{red}{\textbf{0.186}} & \textcolor{blue}{\textbf{0.271}} & 0.195 & 0.280 & 0.175 & 0.265 \\
        \hline
        Wins    & 1 & 1 & 0 & 0 & 3 & 2 & 1 & 0 & 3 & 7 & 0 & 0 & - & - \\
         \hline
         
        \multicolumn{15}{c}{S2IP} \\
        \hline
         \multirow{2}{*}{Dataset} & \multicolumn{2}{c|}{LLM} &  \multicolumn{2}{c|}{Random} & \multicolumn{2}{c|}{LN} & \multicolumn{2}{c|}{Att} & \multicolumn{2}{c|}{Trans} & \multicolumn{2}{c|}{NoLLM}  & \multicolumn{2}{c}{Original} \\
        & MSE & MAE & MSE & MAE & MSE & MAE & MSE & MAE & MSE & MAE & MSE & MAE & MSE & MAE \\
         \hline
        ETTh1    & 0.422 & 0.442 & 0.470 & 0.464 & \textcolor{red}{\textbf{0.410}} & \textcolor{blue}{\textbf{0.435}} & 0.412 & 0.4355 & 0.424 & 0.442 & 0.458 & 0.458 & 0.406 & 0.427 \\
        ETTh2    & 0.366 & 0.403 & 0.376 & 0.412 & \textcolor{red}{\textbf{0.362}} & 0.402 & 0.363 & \textcolor{blue}{\textbf{0.402}} & 0.363 & 0.404 & 0.383 & 0.415 & 0.347 & 0.391  \\
        ETTm1    & 0.351 & 0.389 & 0.347 & \textcolor{blue}{\textbf{0.386}} & \textcolor{red}{\textbf{0.347}} & 0.383 & 0.355 & 0.393 & 0.355 & 0.394 & 0.350 & 0.387 & 0.343 & 0.379  \\
        ETTm2    & 0.273 & 0.331 & 0.271 & 0.329 & 0.270 & 0.330 & 0.272 & 0.329 & \textcolor{red}{\textbf{0.267}} & \textcolor{blue}{\textbf{0.327}} & 0.273 & 0.331 & 0.257 & 0.319  \\
        Illness  & 2.3874 & 1.052 & 2.263 & 1.006 & 2.234 & 0.989 & \textcolor{red}{\textbf{2.015}} & \textcolor{blue}{\textbf{0.927}} & 2.150 & 0.992 & 2.056 & 0.958 & - & - \\
        Weather  & 0.232 & 0.270 & \textcolor{red}{\textbf{0.230}} & \textcolor{blue}{\textbf{0.269}} & 0.230 & 0.270 & 0.232 & 0.273 & 0.231 & 0.269 & 0.238 & 0.274 & 0.222 & 0.259  \\
        Traffic  & \textcolor{red}{\textbf{0.400}} & 0.280 & 0.402 & \textcolor{blue}{\textbf{0.280}} & 0.489 & 0.351 & 0.460 & 0.326 & 0.502 & 0.351 & 0.490 & 0.346 & 0.405 & 0.286  \\
        ECL      & 0.163 & 0.257 & 0.167 & 0.261 & \textcolor{red}{\textbf{0.155}} & \textcolor{blue}{\textbf{0.250}} & 0.165 & 0.259 & 0.168 & 0.262 & 0.177 & 0.270 & 0.161 & 0.257  \\

        \hline
        Wins  & 1 & 0 & 1 & 3 & 4 & 2 & 1 & 2 & 1 & 1 & 0 & 0 & - & - \\
         \hline
    \end{tabular}
    }
    \caption{Long-Term Forecasting Task: All the results are averaged from 4 different prediction length, {24, 36, 48, 60} for Illness and {96, 192, 336, 720} for the others. We highlight the best result in MSE by \textcolor{red}{\textbf{Red}} and MAE by \textcolor{blue}{\textbf{Blue}}. Appendix \ref{sec:experiments_results} shows full experiment results}
    \label{tab:Long-Term Forecasting Results}
\end{table*}

\begin{table*}[ht]
    \centering
    \scalebox{0.9}{
    \begin{tabular}{c|cc|cc|cc|cc|cc|cc|cc}
         \hline
         \multicolumn{15}{c}{One-Fits-ALL} \\
         \hline
         \multirow{2}{*}{Dataset} & \multicolumn{2}{c|}{LLM} &  \multicolumn{2}{c|}{Random} & \multicolumn{2}{c|}{LN} & \multicolumn{2}{c|}{Att} & \multicolumn{2}{c|}{Trans} & \multicolumn{2}{c|}{NoLLM} & \multicolumn{2}{c}{Original} \\
        & MSE & MAE & MSE & MAE & MSE & MAE & MSE & MAE & MSE & MAE & MSE & MAE & MSE & MAE \\
         \hline
        ECL & \textcolor{red}{\textbf{0.085}} & \textcolor{blue}{\textbf{0.204}} & 0.087 & 0.204 & 0.091 & 0.213 & 0.088 & 0.209 & 0.086 & 0.206 & 0.098 & 0.222 & 0.090 & 0.207 \\
        ETTh1       & 0.068 & 0.174 & \textcolor{red}{\textbf{0.067}} & \textcolor{blue}{\textbf{0.172}} & 0.132 & 0.236 & 0.091 & 0.200 & 0.115 & 0.221 & 0.134 & 0.237 & 0.069 & 0.173 \\
        ETTh2       & 0.051 & 0.144 & \textcolor{red}{\textbf{0.050}} & \textcolor{blue}{\textbf{0.143}} & 0.065 & 0.169 & 0.056 & 0.160 & 0.059 & 0.161 & 0.064 & 0.168 & 0.048 & 0.141 \\
        ETTm1       & \textcolor{red}{\textbf{0.027}} & \textcolor{blue}{\textbf{0.108}} & 0.042 & 0.130 & 0.082 & 0.181 & 0.064 & 0.164 & 0.069 & 0.168 & 0.083 & 0.182 & 0.028 & 0.105 \\
        ETTm2       & \textcolor{red}{\textbf{0.022}} & \textcolor{blue}{\textbf{0.086}} & 0.026 & 0.098 & 0.035 & 0.118 & 0.031 & 0.113 & 0.031 & 0.113 & 0.034 & 0.116 & 0.021 & 0.084 \\
        Weather     & 0.032 & 0.058 & \textcolor{red}{\textbf{0.031}} & \textcolor{blue}{\textbf{0.057}} & 0.035 & 0.067 & 0.034 & 0.066 & 0.033 & 0.064 & 0.036 & 0.069 & 0.031 & 0.056 \\
        \hline
        Wins     & 3 & 3 & 3 & 3 & 0 & 0 & 0 & 0 & 0 & 0 & 0 & 0 & - & - \\
         \hline

         \multicolumn{15}{c}{Time-LLM} \\
         \hline
         \multirow{2}{*}{Dataset} & \multicolumn{2}{c|}{LLM} &  \multicolumn{2}{c|}{Random} & \multicolumn{2}{c|}{LN} & \multicolumn{2}{c|}{Att} & \multicolumn{2}{c|}{Trans} & \multicolumn{2}{c|}{NoLLM} & \multicolumn{2}{c}{Original} \\
        & MSE & MAE & MSE & MAE & MSE & MAE & MSE & MAE & MSE & MAE & MSE & MAE & MSE & MAE \\
         \hline
        ECL & 0.009 & 0.082 & \textcolor{red}{\textbf{0.008}} & \textcolor{blue}{\textbf{0.080}} & 0.022 & 0.123 & 0.020 & 0.116 & 0.032 & 0.133 & 0.021 & 0.119 & - & - \\
        ETTh1       & 0.019 & 0.093 & \textcolor{red}{\textbf{0.010}} & \textcolor{blue}{\textbf{0.067}} & 0.063 & 0.168 & 0.054 & 0.151 & 0.011 & 0.070 & \textcolor{red}{\textbf{0.010}} & \textcolor{blue}{\textbf{0.067}} & - & - \\
        ETTh2       & 0.032 & 0.141 & 0.006 & 0.051 & 0.036 & 0.127 & 0.077 & 0.182 & 0.058 & 0.160 & \textcolor{red}{\textbf{0.003}} & \textcolor{blue}{\textbf{0.038}} & - & - \\
        ETTm1       & 0.011 & 0.067 & \textcolor{red}{\textbf{0.006}} & \textcolor{blue}{\textbf{0.053}} & 0.023 & 0.097 & 0.109 & 0.212 & 0.056 & 0.155 & 0.011 & 0.069 & - & - \\
        ETTm2       & 0.012 & 0.071 & \textcolor{red}{\textbf{0.001}} & \textcolor{blue}{\textbf{0.027}} & 0.015 & 0.078 & 0.018 & 0.081 & 0.006 & 0.049 & 0.002 & 0.029 & - & - \\
        Weather     & 0.056 & 0.129 & 0.014 & 0.034 & 0.013 & 0.037 & 0.020 & 0.051 & 0.013 & 0.036 & \textcolor{red}{\textbf{0.008}} & \textcolor{blue}{\textbf{0.030}} & - & - \\
        \hline
        Wins     & 0 & 0 & 4 & 4 & 0 & 0 & 0 & 0 & 0 & 0 & 3 & 3 & - & - \\
         \hline
         
        \multicolumn{15}{c}{CALF} \\
        \hline
         \multirow{2}{*}{Dataset} & \multicolumn{2}{c|}{LLM} &  \multicolumn{2}{c|}{Random} & \multicolumn{2}{c|}{LN} & \multicolumn{2}{c|}{Att} & \multicolumn{2}{c|}{Trans} & \multicolumn{2}{c|}{NoLLM} & \multicolumn{2}{c}{Original} \\
        & MSE & MAE & MSE & MAE & MSE & MAE & MSE & MAE & MSE & MAE & MSE & MAE & MSE & MAE\\
         \hline
        ECL & 0.138 & 0.255 & 0.199 & 0.309 & \textcolor{red}{\textbf{0.104}} & \textcolor{blue}{\textbf{0.225}} & 0.155 & 0.271 & 0.121 & 0.242 & 0.161 & 0.271 & - & - \\
        ETTh1       & 0.299 & 0.306 & 0.363 & 0.363 & \textcolor{red}{\textbf{0.267}} & \textcolor{blue}{\textbf{0.304}} & 0.373 & 0.389 & 0.288 & 0.320 & 0.421 & 0.388 & - & - \\
        ETTh2       & 0.288 & 0.389 & 0.234 & 0.344 & \textcolor{red}{\textbf{0.202}} & \textcolor{blue}{\textbf{0.318}} & 0.205 & 0.320 & 0.388 & 0.402 & 0.231 & 0.342 & - & - \\
        ETTm1       & 0.233 & 0.305 & 0.246 & 0.314 & \textcolor{red}{\textbf{0.160}} & 0.235 & 0.261 & 0.306 & 0.167 & 0.239 & 0.186 & \textcolor{blue}{\textbf{0.215}} & - & - \\
        ETTm2       & 0.171 & 0.291 & 0.169 & 0.308 & 0.105 & 0.225 & 0.133 & 0.257 & \textcolor{red}{\textbf{0.102}} & \textcolor{blue}{\textbf{0.223}} & 0.132 & 0.256 & - & - \\
        Weather     & 0.094 & 0.137 & 0.091 & 0.135 & \textcolor{red}{\textbf{0.085}} & \textcolor{blue}{\textbf{0.134}} & 0.087 & 0.137 & 0.087 & 0.152 & 0.086 & 0.149 & - & - \\
        \hline
        Wins     & 0 & 0 & 0 & 0 & 5 & 4 & 0 & 0 & 1 & 1 & 0 & 1 & - & - \\
         \hline
         
        \multicolumn{15}{c}{S2IP} \\
        \hline
         \multirow{2}{*}{Dataset} & \multicolumn{2}{c|}{LLM} &  \multicolumn{2}{c|}{Random} & \multicolumn{2}{c|}{LN} & \multicolumn{2}{c|}{Att} & \multicolumn{2}{c|}{Trans} & \multicolumn{2}{c|}{NoLLM} & \multicolumn{2}{c}{Original} \\
        & MSE & MAE & MSE & MAE & MSE & MAE & MSE & MAE & MSE & MAE & MSE & MAE & MSE & MAE\\
         \hline
        ECL & 0.083 & 0.204 & 0.081 & 0.202 & 0.078 & 0.205 & 0.071 & 0.191 & 0.071 & 0.193 & \textcolor{red}{\textbf{0.067}} & \textcolor{blue}{\textbf{0.182}} & - & - \\
        ETTh1       & 0.049 & 0.143 & 0.045 & 0.138 & 0.050 & 0.144 & 0.269 & 0.306 & \textcolor{red}{\textbf{0.019}} & \textcolor{blue}{\textbf{0.088}} & 0.125 & 0.233 & - & - \\
        ETTh2       & 0.028 & 0.110 & 0.126 & 0.233 & \textcolor{red}{\textbf{0.015}} & \textcolor{blue}{\textbf{0.083}} & 0.890 & 0.554 & 0.085 & 0.194 & 0.066 & 0.168 & - & - \\
        ETTm1       & 0.025 & 0.104 & \textcolor{red}{\textbf{0.012}} & \textcolor{blue}{\textbf{0.073}} & 0.022 & 0.099 & 0.071 & 0.163 & 0.072 & 0.172 & 0.147 & 0.260 & - & - \\
        ETTm2       & 0.005 & 0.048 & \textcolor{red}{\textbf{0.004}} & \textcolor{blue}{\textbf{0.042}} & 0.008 & 0.058 & 0.047 & 0.141 & 0.044 & 0.136 & 0.049 & 0.157 & - & - \\
        Weather     & 0.012 & 0.051 & 0.018 & 0.080 & \textcolor{red}{\textbf{0.010}} & \textcolor{blue}{\textbf{0.058}} & 0.052 & 0.105 & 0.040 & 0.080 & 0.111 & 0.169 & - & - \\
        \hline
        Wins        & 0 & 0 & 2 & 2 & 2 & 2 & 0 & 0 & 1 & 1 & 1 & 1 & - & - \\
         \hline
    \end{tabular}
    }
    \caption{Imputation Task: All results are averaged from experiments with missing data proportion \{ 12.5\%, 25\%, 37.5\%, 50\% \}. We highlight the best performer in MSE index by \textcolor{red}{\textbf{Red}} and MAE by \textcolor{blue}{\textbf{Blue}}. Appendix \ref{sec:experiments_results} shows full experiments results}
    \label{tab:Time Series Imputation Results}
\end{table*}

\begin{table}
    \centering
    \scalebox{0.8}{
    \begin{tabular}{c|c|cccccc}
         \multirow{2}{*}{Method} & \multirow{2}{*}{Variants} & \multicolumn{6}{c}{Dataset} \\
           & & SMD & MSL & SMAP & SWaT & PSM & Win \\
         \hline
          \multirow{6}{*}{\begin{turn}{90} 
         OFA 
         \end{turn}} 
         & LLM      & 84.23 & \textcolor{red}{\textbf{82.43}} & 68.88 &  92.58  & 97.08 & 1 \\
         & Random   & 84.34 & 81.17 & \textcolor{red}{\textbf{69.22}} & 92.51 & 97.08 & 1 \\
         & Linear   & \textcolor{red}{\textbf{86.78}} & 77.50 & 66.06 & 92.45 & \textcolor{red}{\textbf{97.20}} & 2 \\
         & Att      & 86.32 & 77.52 & 68.91 & \textcolor{red}{\textbf{92.84}} & 94.97 & 1 \\
         & Trans    & 85.93 & 77.90 & 67.08 & 88.75 & 96.80  & 0 \\
         & NoLLM  & 83.89 & 82.00 & 66.12 & 91.63 & 94.72  & 0 \\
         & Original & 86.89 & 82.45 & 68.88 & 94.23 & 97.13 & - \\
         \hline
        
          \multirow{6}{*}{\begin{turn}{90} 
         Time-LLM 
         \end{turn}} 
         & LLM      & 82.13 & 66.74 & 66.55 & 86.26 & 88.01 & 0 \\
         & Random   & 82.22 & 66.29 & \textcolor{red}{\textbf{66.88}} & \textcolor{red}{\textbf{87.11}} & 85.68 & 2 \\
         & Linear   & \textcolor{red}{\textbf{82.65}} & 66.40 & 65.62 & 84.54 & 88.28 & 1 \\
         & Att      & 80.50 & 56.70 & 65.62 & 82.57 & 84.96 & 0 \\
         & Trans    & 81.18 & \textcolor{red}{\textbf{65.40}} & 65.60 & 84.56 & \textcolor{red}{\textbf{88.56}} & 2 \\
         & NoLLM  & 82.05 & 63.33 & 65.92 & 82.82 & 86.03 & 0 \\
         & Original & - & - & - & - & - & - \\
         \hline

          \multirow{6}{*}{\begin{turn}{90} 
         CALF 
         \end{turn}} 
         & LLM     & 83.14 & \textcolor{red}{\textbf{83.04}} & 67.44 & \textcolor{red}{\textbf{94.13}} & 94.75 & 2 \\
         & Random  & 83.50 & 83.02 & 67.40 & \textcolor{red}{\textbf{94.13}} & 96.63 & 1 \\
         & Linear  & 84.31 & 74.56 & 66.19 & 91.02 & 94.58 & 0 \\
         & Att    & 83.68 & 83.04 & \textcolor{red}{\textbf{68.87}} & 87.02 & \textcolor{red}{\textbf{96.76}} & 2 \\
         & Trans   & \textcolor{red}{\textbf{84.41}} & 72.64 & 66.27 & 87.24 & 95.07 & 1 \\
         & NoLLM & 82.82 & \textcolor{red}{\textbf{83.04}} & 67.32 & \textcolor{red}{\textbf{94.13}} & 96.70 & 2 \\
         & Original & - & - & - & - & - & - \\
         \hline

          \multirow{6}{*}{\begin{turn}{90} 
         S2IP 
         \end{turn}} 
         & LLM     & 81.17 & 77.93 & 68.94 & 81.45 & 95.17 & 0 \\
         & Random  & 81.44 & 77.98 & 68.10 & \textcolor{red}{\textbf{81.46}} & 95.44 & 1 \\
         & Linear  & \textcolor{red}{\textbf{82.99}} & 78.11 & \textcolor{red}{\textbf{70.78}} & 79.23 & 96.45 & 2 \\
         & Att     & 80.35 & 68.52 & 67.80 & 80.08 & \textcolor{red}{\textbf{96.85}} & 1 \\
         & Trans   & 80.66 & 69.04 & 66.31 & 80.94 & 96.29 & 0 \\
         & NoLLM & 81.28 & \textcolor{red}{\textbf{79.45}} & 68.83 & 79.24 & 96.14 & 1 \\
         & Original & - & - & - & - & - & - \\
         \hline
    \end{tabular}
    }
    \caption{Anomaly Detection Task: We calculate the F1 score (\%) for each experiment and highlight the best performance compared in 6 different models by \textcolor{red}{\textbf{Red}}. Appendix \ref{sec:experiments_results} shows full experiment results for Precision and Recall}
    \label{tab:anomaly_detection_results}
\end{table}

\begin{table}[ht]
    \centering
    \scalebox{0.88}{
    \begin{tabular}{c|c|c|c|c|c|c}
         \hline
         Method & LLM &  Random & LN & Att & Trans & NoLLM \\
         \hline
         OFA      & 71.82 & 72.21 & 71.06 & \textcolor{red}{\textbf{72.25}} & 71.77 & 72.18 \\
         Time-LLM & 37.68 & 37.38 & 37.32 & 35.24 & 36.63 & \textcolor{red}{\textbf{40.83}} \\
         CALF     & 69.78 & 69.46 & 70.19 & \textcolor{red}{\textbf{70.43}} & 70.29 & 69.17 \\
         S2IP     & 68.96 & 68.73 & 67.65 & 69.00 & 67.71 & \textcolor{red}{\textbf{69.15}} \\
         \hline
    \end{tabular}
    }
    \caption{Multivariate Classification Task: We calculate the Accuracy (\%) by averaging the results from 10 classification dataset. We highlight the best results by \textcolor{red}{\textbf{Red}}. Appendix \ref{sec:experiments_results} shows full experiment results.}
    \label{tab:multivariate_results}
\end{table}

\section{Results and Discussion}
\subsection{Is LLM Improving Performance for Time Series Modeling?}
To validate the effectiveness of Language knowledge, we implement intensive ablation study in four significant tasks. As Time-LLM, CALF, S2IP are specific forecasting models without implementation for other tasks. We use the benchmark implementation from Time Series Library \cite{wang2024tssurvey} to preprocess data and change only the projection head for the downstream task. To be a fair comparison, we keep the original implementation and parameters given by the authors and we search for the best learning rate for model convergence for other tasks.

\textbf{Forecasting}: 
Table. \ref{tab:Long-Term Forecasting Results} shows the average forecasting results. Impressively, the Random variant achieves similar performance to, and even outperforms, the original LLM on Time-LLM and S2IP across different scales of forecasting datasets. Among all models and variants, the single-layer linear variant consistently achieves the best results across the majority of forecasting tasks. This finding aligns with the nature of time series data, where it's sparsity and low information density often lead to overfitting by employing large models \cite{zeng2023transformers, xu2023fits}. Notably, the reproduced results of Time-LLM are significantly lower than the original reported performance, showing no wins when compared to simpler baselines. Although Time-LLM utilizes cross-attention and in-context learning to align time series data with the language tokens, the Random variant outperforms LLMs across all datasets. This suggests that the language knowledge embedded in the 32-layer LLaMa model used by Time-LLM may not be fully activated. Interestingly, we observed that the use of LLMs does not consistently outperform performance and even degrade it for datasets such as ETTh1, ETTh2, ETTm1, ETTm2, and Illness compared to the original LLM implementation. This also implies that LLMs may be prone to overfitting in time series forecasting tasks. For larger datasets, such as Weather, Traffic, and ECL, LLM models still fail to perform well. Although LLM models like OFA, CALF, and S2IP achieve some victories across various datasets, the improvements are not substantial when compared to other baseline models. In short, the employment of LLMs does not show a strong performance discrepancy, and even degrade the performance over simple baseline in time series forecasting task.

\textbf{Imputation}: 
Table \ref{tab:Time Series Imputation Results} presents the imputation performance across various models. Notably, OFA outperforms others, winning in 6 out of 12 cases across all datasets, though its improvement over the Random variant is not substantial. Interestingly, the LLM variants of TimeLLM, CALF, and S2IP did not achieve any wins in terms of MSE or MAE metrics. However, the Random variant of TimeLLM and the LayerNorm variant of CALF show significant improvements of 58\% and 24\%, respectively, over their LLM counterparts. Among the LLM-based models, CALF demonstrates the poorest performance in computational tasks, falling well behind other reprogramming methods. This suggests challenges in aligning time series predictions with language-based learning. In contrast, TimeLLM performs impressively within the LLM4TS approaches, showing substantial improvements, particularly on smaller datasets such as ETTh1 and ETTh2. OFA, despite being the simplest reprogramming approach, fine-tuning only the LayerNorm layer, achieves the highest number of wins compared to other LLM4TS methods. Overall, this analysis indicates that LLM backbones may not be well-suited for imputation tasks, as several simpler baseline methods outperform them in many cases.

\textbf{Anomaly Detection}:
In Table. \ref{tab:anomaly_detection_results}, we observe the same conclusion as forecasting, that LLM is not the key component for anomaly detection. Anomaly Detection task has imbalanced distribution label, we reported the F1 score in Table \ref{tab:anomaly_detection_results} and compare four LLM4TS methods with their simple variants. We also include the additional results in Appendix \ref{sec:experiments_results} for details comparison. OFA with LLM wins only MSL dataset and the performance does not significantly outperform other simple variants. Time-LLM and S2IP perform much worse than OFA and has zero wins compared to other ablation model. CALF wins two dataset and the performance is similar to NoLLM, making LLM is not computational worthy for this minor improvement. In short, the best backbone across all dataset and method is Linear and Random with 5/20 wins and the LLM wins only 3/20.

\textbf{Classification}: 
In Table. \ref{tab:multivariate_results}, we follow the same experimental setup as in OFA, selecting 10 multivariate datasets from the UEA time series archive. The average results from these datasets are presented in Table \ref{tab:multi_classification}. Notably, although the LLM model does not outperform all of the simple baseline models, it shows significant improvement despite being among the lower-performing models overall. Moreover, we observed that Time-LLM is not well-suited for multivariate classification tasks without substantial modifications to the original implementation, as it is primarily designed to generate prefix prompts for univariate time series. As a result, Time-LLM often fails to capture interactions between different time series variables, while also incurring extremely high computational costs per sample. We give full experiments results in Appendix.\ref{sec:experiments_results}.

Overall, our extensive experiments show the existing reprogramming approach usually fails on time series modelling while the employment of LLM significantly increase the computational efforts. The performance of reprogramming LLM is less effective than simple linear and the impressive results from Random ablation further demonstrated the performance of LLM4TS does not rely on the language knowledge but time series data itself.

\subsection{Can LLM Effectively Capture Temporal Dependency?}
Unlike language data, time series data inherently exhibits a stronger temporal relationship due to its characteristic of decaying auto-correlation over time\cite{nie2023timeseriesworth64,hamilton2020time}. Given that language models are typically not trained on time series data, it raises questions about their ability to recognise and learn the temporal dependencies inherent in time series. This difference suggests that the temporal dynamics in time series may not be fully captured by language models trained exclusively on linguistic patterns. In time series analysis, one common assumption is that a forecasting procedure is successful and sufficient if the prediction residual is independent \cite{hamilton2020time,shumway2000time}. Then, it is intuitive that the correlation of residual indicates the dependency of model prediction to label. A model can be claimed as "Capturing Temporal Relationship" if the residual of each time step is independent to each other. Therefore, we take the long-term forecasting task to empirically analyze residual independence to interpret whether LLM can capture the temporal relationship. We introduce Durbin–Watson Test \cite{durbin1971testing} to quantitatively interpret the residual independence as shown in Equation. \ref{eq:durbin-watson} and residual ACF plots to visualize the ability to capture temporal features. Given the residual between prediction and true label $e_t = {e_1, \dots, e_t}_{t=1}^L$, where $L$ is the sequence length of time series data. Durbin–Watson statistic is approximately equal to 2 when the residual correlation is independent (see Proof in Appendix \ref{sec:proof_Durbin}), $\mathcal{D}>2$ is negatively correlated and $\mathcal{D}<2$ is positively correlated.

\begin{figure*}
    \centering
    \includegraphics[width=0.95\linewidth]{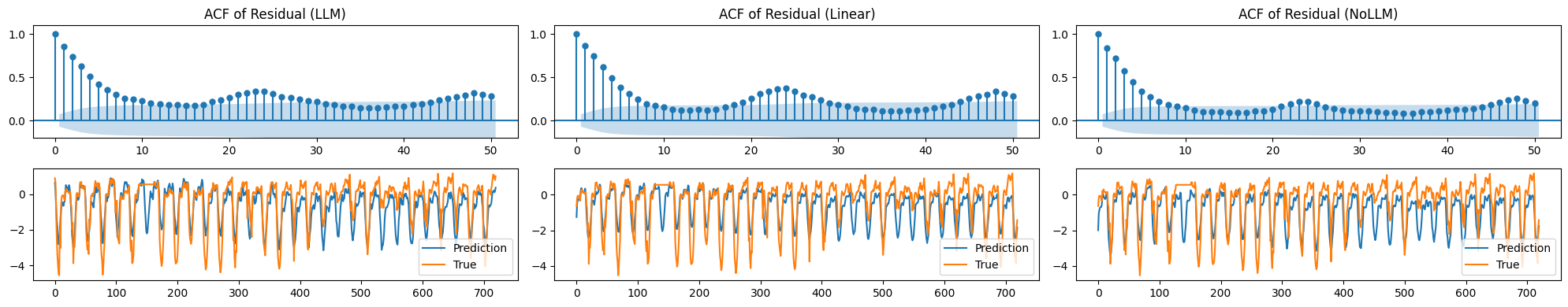}
    \caption{Residual ACF and forecasting plots of OFA on ETTh1 dataset. The Durbin-Watson Statistic for LLM, Linear and NoLLM are 0.3210, 0.3383, 0.3325}
    \label{fig:OFA_ETTh1_acf_pacf}
\end{figure*}

\begin{figure*}
    \centering
    \includegraphics[width=0.95\linewidth]{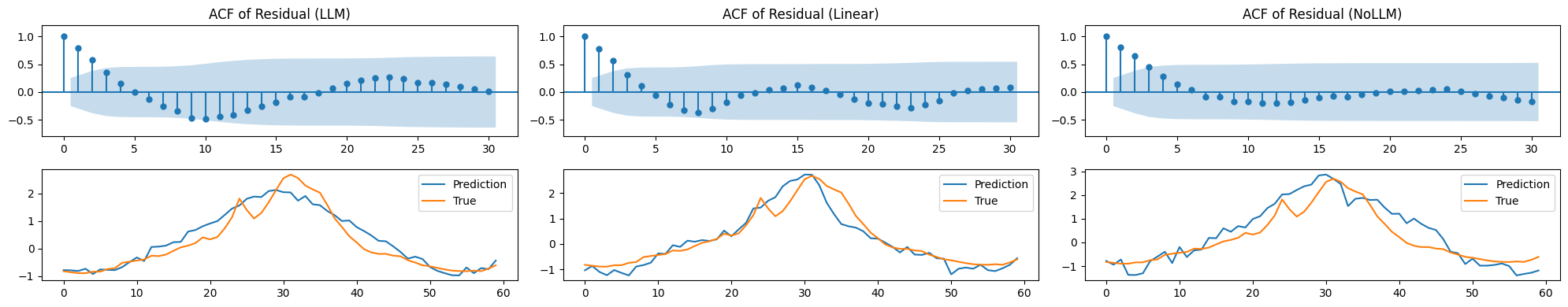}
    \caption{Residual ACF and forecasting plots of OFA on Illness dataset. The Average Durbin-Watson Statistic for LLM, Linear and NoLLM are 0.127, 0.170, 0.167}
    \label{fig:OFA_Ill_acf_pacf}
\end{figure*}

\begin{figure*}
    \centering
    \includegraphics[width=0.95\linewidth]{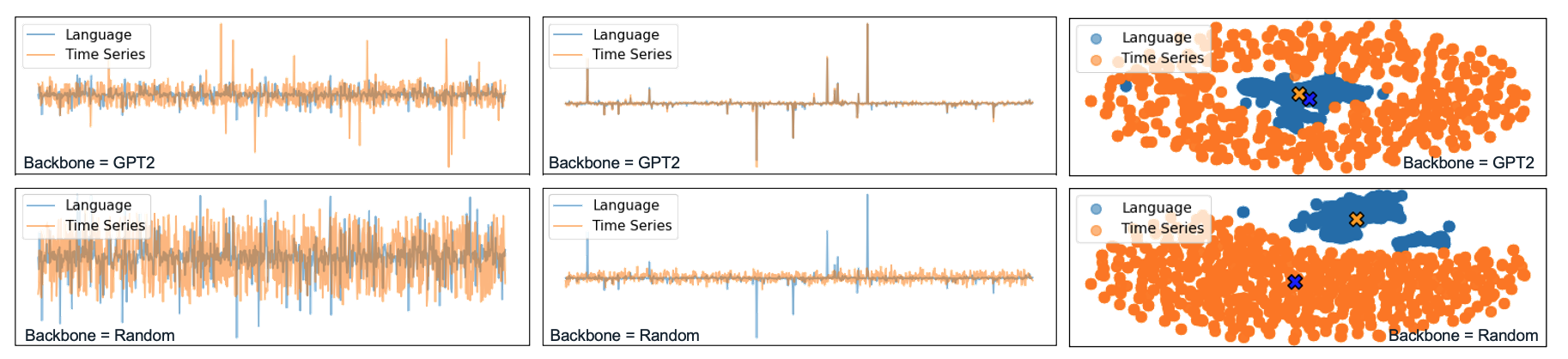}
    \caption{OFA on ETTh1 Dataset: (Column 1): Feature Map before Backbone. (Column 2): Feature Map after Backbone, GPT2 Backbone displays a "pseudo-alignment" behaviour. (Column 3): UMAP Plot of Random and LLM Backbone, "Pseudo-alignment" simply transfer the centroid of TS token without interactions between language and TS.}
    \label{fig:TS-TXT_Distribution}
\end{figure*}

\begin{equation}
    \mathcal{D} = \frac{\sum_{t=2}^L (e_t - e_{t-1})^2)}{\sum_{t=1}^L e_t^2} \label{eq:durbin-watson}
\end{equation}

Figure \ref{fig:OFA_ETTh1_acf_pacf} and Figure \ref{fig:OFA_Ill_acf_pacf} display the ACF and forecasting plots for the OFA model. Notably, the GPT-2 performs statistically worse than both the Linear and NoLLM models, demonstrating a reduced ability to capture temporal features, as reflected by diverged Durbin-Watson statistics away from 2. Moreover, the residual ACF plots for the LLM across both datasets exhibit clear sinusoidal patterns, typically indicative of strong seasonality and correlation patterns in the residuals, which violates the assumption of residual independence in time series modelling. In the ETTh1 dataset, this sinusoidal trend is particularly significant than Linear and NoLLM, extending beyond the confidence intervals. Conversely, for both the Linear and NoLLM models, most of the correlation scores fall within the confidence interval, suggesting no significant autocorrelation at different lags underneath the confidence region. This indicates that these models have effectively captured temporal feature of the time series in the Illness dataset, leaving little insignificant variance. Although the LLM’s results for the Illness dataset appear to meet the residual independence assumption, the autocorrelation still displays a more pronounced sinusoidal pattern compared to the Linear and NoLLM models. LLMs even sometimes displays better MSE performance, but the Durbin-Watson statistic is lower than the simple baseline. For example, CALF shows lower MSE and MAE in ETTh1 than Linear and NoLLM variants, but their corresponding Durbin-Statistics are 0.3450, 0.3588, 0.3641 for LLMs, Linear and NoLLM respectively. The additional results, presented in the Appendix \ref{sec:Additional_Visualization}, further support these findings.

\subsection{Is Reprogramming Actually Aligning Time Series to Language?}
For a successful reprogramming stated in Theorem 3.1, the distribution of time series data $t_i$ and language data $s_i$ should be aligned together to leverage the language knowledge of LLM. Our experiments indicate LLM and Random initialization have little gaps in performance, motivating us to investigate whether the time and language modality are actually aligned in a common space.

Instead of using the probabilistic dimension reduction approach such as T-SNE and PCA, we employ UMAP\cite{mcinnes2018umap} manifold projection to plot 2D data representation with 10 neighbours, which seeks to preserve both local and global manifold structure by constructing a high-dimensional graph of the data and optimizing a low-dimensional projection that preserves the graph structure. An ideal alignment of two modalities should display coherent representation, and the time series should display a similar internal structure as language before and after LLM backbone. Moreover, the embedding structure of the time series from LLM should display distinct discrepancy from the Random initialization model, if language knowledge is actually interact with time series, since Random initialization preserves only noise to input time series.

As shown in Figure. \ref{fig:TS-TXT_Distribution}, we find that fine-tuning LayerNorm layer with pre-trained LLM only maps the centroid of time series data to $\mathcal{N} (0, 1)$ instead of actual alignment in token level, where time series modality displays higher variance before and after LLM. This leads to high similarity in mean feature maps after passing through the LLM (row 1). The identical observation is also found in S2IP and TimeLLM (See Figures in Appendix \ref{sec:Additional_Visualization}). This is because the pre-trained multi-head attention performs a role to re-distribute the feature into a format tailored for language data \cite{voita2019analyzing}. Multihead attention mechanisms, when trained on language data, emphasizes specific key features relevant to language, indicating the highlight feature value in language token. However, applying the same weights directly to time series data might not be optimal. It can potentially accentuate less significant features, which may contribute minimally to downstream tasks in time series applications. Thus, while LLM attempts to mimic the feature distribution of language, it may inadvertently disrupt the semantic integrity of the time series data, given that the multi-head attention mechanism is optimized for language, not time series. Moreover, it is intuitive to understand that the language tokes were trained by unsupervised manner from one-hot embedding. The normalization of embedding layer is well-initialized in Gaussian $N(0, 1)$ while the actual reading of time series data does not strictly follow $N(0, 1)$, leading to large variance feature distribution. LayerNorm layer can transform each feature to $N(0, 1)$, maintaining the internal structure of the original data rather than aligning time series directly to language. In contrast, random variants—where the multi-head attention and feed-forward layers are randomly initialized and only train the LayerNorm layer—show a more varied feature distribution than LLM due to randomly initialized attention layer. It is also easy to understand that the UMAP plot of Random initialization shows distinct discrepancy with language tokens since the language knowledge is removed. Considering the performance of Random and LLM are similar across 4 time series task, we observe that the internal structure of time series exhibit similar in LLM and Random (See Figure. \ref{fig:LLM_Random_UMAP_Comparison} in Appendix \ref{sec:Additional_Visualization}), which further prove our stands that multihead attention trained on language knowledge can potentially raise those irrelevant feature in time series.

\begin{table*}[ht]
    \centering
    \resizebox{0.9\linewidth}{!}{%
    \begin{tabular}{cc|cc|cc|cc|cc|cc|cc|cc|cc}
        \hline
         \multicolumn{2}{c|}{Method} & \multicolumn{2}{c|}{OFA} &  \multicolumn{2}{c|}{TimeLLM} & \multicolumn{2}{c|}{CAFL} & \multicolumn{2}{c|}{S2IP} & \multicolumn{2}{c|}{OFA w/ M} &  \multicolumn{2}{c|}{TimeLLM w/ M} & \multicolumn{2}{c|}{CAFL w/ M} & \multicolumn{2}{c}{S2IP w/ M} \\
         \multicolumn{2}{c|}{Pred Length} & MSE & MAE & MSE & MAE & MSE & MAE & MSE & MAE & MSE & MAE & MSE & MAE & MSE & MAE & MSE & MAE \\
         \hline
        \multirow{5}{*}{\begin{turn}{90} 
         ETTh1 
         \end{turn}} 
         & 96      & 0.392 & 0.404 & 0.400 & 0.400 & 0.377 & 0.394 & 0.380 & 0.407 & 0.380 & 0.404 & 0.380 & 0.399 & 0.386 & 0.397 & 0.384 & 0.413 \\
         & 192     & 0.432 & 0.430 & 0.424 & 0.434 & 0.428 & 0.425 & 0.407 & 0.425 & 0.420 & 0.427 & 0.416 & 0.436 & 0.427 & 0.422 & 0.408 & 0.428 \\
         & 336     & 0.441 & 0.435 & 0.435 & 0.448 & 0.477 & 0.450 & 0.433 & 0.450 & 0.433 & 0.436 & 0.428 & 0.443 & 0.473 & 0.449 & 0.422 & 0.442 \\
         & 720     & 0.468 & 0.471 & 0.447 & 0.474 & 0.473 & 0.466 & 0.467 & 0.483 & 0.437 & 0.457 & 0.439 & 0.464 & 0.468 & 0.465 & 0.457 & 0.474 \\
         & Average & 0.433 & 0.435 & 0.426 & 0.439 & 0.439 & 0.434 & 0.422 & 0.442 & 0.418 & 0.431 & 0.416 & 0.435 & 0.439 & 0.433 & 0.418 & 0.439 \\
        \hline
        \multirow{5}{*}{\begin{turn}{90} 
         Illness 
         \end{turn}} 
         & 24      & 2.205 & 0.982 & 1.938 & 0.898 & 2.271 & 1.008 & 2.622 & 1.113 & 2.060 & 0.945 & 1.991 & 0.871 & 2.226 & 0.961 & 2.136 & 0.979 \\
         & 36      & 2.166 & 0.969 & 1.812 & 0.887 & 2.365 & 1.029 & 2.515 & 1.098 & 1.946 & 0.934 & 1.953 & 0.879 & 1.831 & 0.848 & 2.183 & 0.990 \\
         & 48      & 2.220 & 0.963 & 1.734 & 0.887 & 2.395 & 1.046 & 2.530 & 1.066 & 2.135 & 0.999 & 1.851 & 0.875 & 1.815 & 0.860 & 2.158 & 0.976 \\
         & 60      & 2.246 & 1.010 & 1.945 & 0.957 & 2.415 & 1.056 & 2.329 & 1.065 & 2.062 & 1.015 & 1.906 & 0.914 & 1.892 & 0.879 & 2.085 & 1.009 \\
         & Average & 2.209 & 0.981 & 1.857 & 0.907 & 2.361 & 1.035 & 2.387 & 1.052 & 2.051 & 0.973 & 1.925 & 0.885 & 1.941 & 0.887 & 2.141 & 0.989 \\
         \hline
    \end{tabular}
    }
    \caption{Preliminary Performance Comparison on Mixer LLM: "w/ M" indicate the model with Mixer module}
    \label{tab:mixer_performance_table}
\end{table*}

\begin{figure}[ht]
    \centering
    \includegraphics[width=1\linewidth]{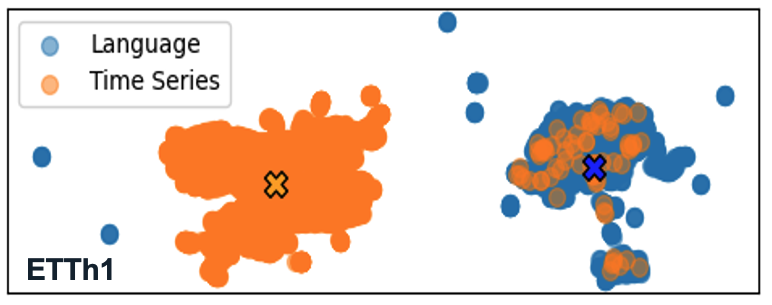}
    \caption{UMAP plot of OFA w/ Mixer in ETTh1 and Illness}
    \label{fig:Mixer_UMAP_plots}
\end{figure}

Therefore, we conclude from our empirical evidence that finetuning LayerNorm in OFA and S2IP, and Cross-Attention in TimeLLM is hard on actually aligning the time series modality to leverage the reasoning ability of LLM and the "pseudo-alignment" can potentially downgrade the performance in downstream task. Furthermore, we plot the UMAP plots to show whether the internal structure remains the same in random and LLM backbone in Figure \ref{fig:LLM_Random_UMAP_Comparison} of Appendix \ref{sec:Additional_Visualization}. We observed that none of methods changes the internal structure of time series data in random initialization and LLM resulting in identical distribution of time series in random intializatio and pre-trained LLM. CALF, on the other hand, finetunes the whole model using LoRA still keep the global structure of time series similar with Random but the distribution is shifted. This solid evidence further supports that \textbf{the performance of LLM4TS method came from the internal structure of time series data itself instead of language knowledge}. The existing reprogramming approaches are difficult to leverage the language knowledge on time series data.

\section{A Preliminary Solution to Pseudo-Alignment}
With respect to the internal structure of time series data, complete alignment may be harmful to downstream performance as language can be difficult to describe temporal relationship of time series. We propose a demo solution to show the feasibility of addressing "Pseudo-Alignment" by utilizing the . We would like to leave the complete solution to future work. This demo solution gains inspiration from MLP-mixer as modality alignment and implicitly fusion module as stated in \cite{tolstikhin2021mlp,xin2025med3dvlm}. Specifically, the idea is to mix-up the semantic embedding from word corpus and time series token by two mixing MLP to form a meta text tokens containing proportion of time series feature so that LLM can understand what pattern is useful. We select TopK semantic text token $T_k, k \in [0, 1, \dots, K]$ extracting from the corpus by a linear mapping and attention mechanism, similar to the text prototype as Time-LLM \cite{jin2023time}. $T_k$ will be concatenated with time series token $\mathcal{X}_t$ by patch embedding. A two-layers mixing MLP $M_1 \in \mathbb{R}^{m \times k}$ applied on token-dimension to shrink the long-sequence of the concatenated sequence and another two-layers mixing MLP $M_2$ is applied on text-dimension to fuse the feature from the combination of text and time. This compression serves as an inexplicit alignment by the linear combination of time and text token across feature and sequence dimension. We find this approach is particularly useful as it consider the feature processing as multi-modal learning setting and align them implicitly without alignment loss function involved. We can see the UMAP plot shown in Figure \ref{fig:Mixer_UMAP_plots}. The time series data token is mapping on the general outline of Language tokens instead of moving the centroid over language distribution. In Table \ref{tab:mixer_performance_table}, we compare the performance of baseline models with Mixer module inserted before LLMs without changing any parameter indicated in the original implementation. It is clear that the performance improvement is notably significant. The mixer we employed in this paper is a preliminary solution to address "Pseudo-Alignment" problem and we consider a comprehensive solution as a future work.

\section{Conclusion}
This work questions the effectiveness of applying Large Language Models to general time series modeling tasks such as forecasting, imputation, anomaly detection, and classification. Our findings demonstrate that language knowledge is not essential for time series analysis, but depress the performance in many time series dataset and significantly increase the computational resource. Furthermore, reprogramming approaches, which aim to adapt LLMs to time series tasks, fail to capture temporal features as effective as a simple models. One of the primary reasons for this shortfall is the "pseudo-alignment" between the inherent structure of time series and the language, which limits the ability of language knowledge to contribute meaningfully to time series tasks.

\section{Acknowledge} 
This work was supported by Australian Research Council ARC Early Career Industry Fellowship (Grant No.IE240100275), Australian Research Council Discovery Projects (Grant No.DP240103070), Australian Research Council Linkage(Grant No.LP230200821),  University ofAdelaide, Global Partnership Engagement Fund, 2025.

\bibliographystyle{ACM-Reference-Format}
\balance
\bibliography{main}


\appendix
\newpage
\section{Proof of Theorem 3.1} \label{sec:proof_theorem}
Denote $t_i$ and $s_i$ as arbitrary time series and language samples. In model reprogramming setting, we hypothesis that the expectation of time series $E_T$ can be represented by language $E_S$. BY the expectation definition, $E_T = E\left[ p(s_i) \right] = \int s_i dp(s_i)$ and $E_S = E\left[ q(Rep(t_i)) \right] = \int Rep(t_i) dq(Rep(t_i))$, where $Rep(\cdot)$ is the reprogramming operation, we denote $Rep(t_i)$ as $t_i^{*}$ for simplicity. In practice, We map the time series and language data into the same feature space and we have an arbitrary function $\gamma \in \Gamma(t_i, s_i)$. Therefore, the key success of model reprogramming is minimize the distance between $E_T$ and $E_S$ in order to fully utilize the knowledge of LLM.

\begin{align}
    | E_S - E_T | 
    &= | \int s_i dp(s_i) - \int t_i^{*} dq(t_i^{*}) | \\
    &= | \int s_i dp(s_i) - \int t_i^{*} dq(t_i^{*})) | \\
    &= | \int s_i - t_i^{*} d\gamma(s_i, t_i^{*}) | \\
    &= \inf_{\gamma \in \Gamma_{s_i,t_i^{*}}} \int |s_i - t_i^{*}| d\gamma(s_i, t_i^{*}) \\
    &= K \mathcal{W}_1(s_i,t_i^{*})
\end{align}

The Kantorovich-Rubinstein Theorem \cite{edwards2011kantorovich} states that

\begin{equation}
    \mathcal{W}_1(\mu,\nu) = \sup_{f\in Lip_1} E_{x \sim \mu}[f(x)] - E_{y \sim \nu}[f(y)]
\end{equation}

where $f(\cdot)$ belongs to a set of 1-Lipschitz functions. However, neural network is not a typically 1-Lipschitz function if the training is not intended to constraint the 1-Lipschit. Generally, $\mathcal{W}_1(\mu,\nu)$ can be represented as:

\begin{equation}
    \mathcal{W}_1(\mu,\nu) = \frac{1}{K} \sup_{f\in Lip_K} E_{x \sim \mu}[f(x)] - E_{y \sim \nu}[f(y)]
\end{equation}

where $f\in Lip_K$ is a K-Lipschitz function.

\section{Proof of Durbin-Watson Statistic} \label{sec:proof_Durbin}
Given the residual between prediction and true label $e_t = {e_1, \dots, e_t}_{t=1}^L$, where $L$ is the sequence length of time series data. We now show that the Durbin-Watson statistic is closed to 2 when the residual correlation is independent. We give Durbin-Watson Statistic here for reference.

\begin{equation}
    \mathcal{D}_w = \frac{\sum_{t=2}^L (e_t - e_{t-1})^2)}{\sum_{t=1}^L e_t^2}
\end{equation}

We start with expanding the numerator $ \sum_{t=2}^L (e_t - e_{t-1})^2 = \sum_{t=2}^Le_t^2 - \sum_{t=2}^L2e_te_{t-1} + \sum_{t=2}^Le_{t-1}^2$. Since $\sum_{t=2}^Le_{t-1}^2$ is just a shift of $\sum_{t=2}^Le_{t}^2$ by one step, we can approximately consider $\sum_{t=2}^Le_{t-1}^2 \simeq \sum_{t=2}^Le_{t}^2$. When there is no autocorrelation, the residual $e_t$ are uncorrelated with $e_{t-1}$, meaning the expectation of their product is zero $E[e_te_{t-1}] = 0$. Therefore, we obtain

\begin{equation}
    \mathcal{D}_w = \frac{2 \sum_{t=2}^L e_t^2 - 2 \sum_{t=2}^L e_te_{t-1}}{\sum_{t=1}^L e_t^2} = \frac{2 \sum_{t=2}^L e_t^2}{\sum_{t=1}^L e_t^2} = 2
\end{equation}

Similarly, $E[e_te_{t-1}] > 0$ leads to positively correlation such that $\mathcal{D}_w < 2$ and $E[e_te_{t-1}] < 0$ leads to negatively correlation such that $\mathcal{D}_w > 2$.

\section{Results Reproduction Note} \label{sec:reproduction note}

\subsection{Model and Parameter}
We reproduce the results of OFA, TimeLLM, CALF and S2IP using a public time series benchmark adopt by the authors: Time Series Library\cite{wang2024tssurvey}. We keep the same data preprocessing methods as the original papers. Since OFA implement four time series task, we do not change any model and dataset parameter of OFA. All LLM4TS methods have implement the time series forecasting task, we keep the original parameters as it is except for batch size in Time-LLM to fit our device. Specifically, we list all the changes we made for all models and tasks:

\begin{itemize}
    \item \textbf{Forecasting}: Time-LLM was originally trained on 8 $\times$ A100 80G with 24 batch size. We reduce the batch size to 6 to avoid OOM problem.
    \item \textbf{Imputation and Anomaly Detection}: We keep the OFA parameter as it is. We change only the output projection head of CALF and S2IP from prediction length to sequence length, which is consistent as the implementation of Time Series Library. We fetch the dataset description from the original dataset paper for the prompts of TimeLLM, which will be offered in our GitHub repository.
    \item \textbf{Classification}: We keep the original implementation for OFA. For CALF, S2IP and TimeLLM, we change only the output projection head to number of classes and patch the time series with number of channel same as OFA. We fetch the dataset description from the original dataset paper for TimeLLM, which will be offered in our GitHub repository.
\end{itemize}

\subsection{Accelerate Data Prepossessing}
In the original implementation, S2IP and TimeLLM generate data in model forward function, which uses only one CPU core to process the batch data and significantly slow down the training speed without fully utilizing GPU capacity of batch operation. Therefore, we implement the data processing procedure of S2IP and Time-LLM in Torch DataLoader class for parallel data processing and only retain GPU parallel computing operation in model forward function. We attach original implementation alongside our optimal code for results reproduction.

\section{Full Experiments Results} \label{sec:experiments_results}
\begin{table*}[ht]
    \centering

\begin{tabular}{cc|cc|cc|cc|cc|cc|cc|cc}
         \hline
         \multicolumn{2}{c|}{Method} & \multicolumn{2}{c|}{LLM} &  \multicolumn{2}{c|}{Random} & \multicolumn{2}{c|}{LN} & \multicolumn{2}{c|}{Att} & \multicolumn{2}{c|}{Trans} & \multicolumn{2}{c|}{NoLLM} & \multicolumn{2}{c}{Original} \\
         \multicolumn{2}{c|}{Pred Length} & MSE & MAE & MSE & MAE & MSE & MAE & MSE & MAE & MSE & MAE & MSE & MAE & MSE & MAE \\
         \hline
         
        \multirow{5}{*}{\begin{turn}{90} 
         ETTh1 
         \end{turn}} 
            & 96  & 0.392 & \textcolor{blue}{\textbf{0.404}} & 0.465 & 0.451 & 0.399 & 0.412 & 0.411 & 0.420 & \textcolor{red}{\textbf{0.381}} & \textcolor{blue}{\textbf{0.404}} & 0.397 & 0.413 & 0.376 & 0.397 \\
            & 192 & 0.432 & 0.430 & 0.505 & 0.468 & \textcolor{red}{\textbf{0.421}} & \textcolor{blue}{\textbf{0.422}} & 0.424 & 0.434 & 0.432 & 0.432 & 0.425 & 0.428 & 0.416 & 0.418 \\
            & 336 & \textcolor{red}{\textbf{0.441}} & \textcolor{blue}{\textbf{0.435}} & 0.523 & 0.482 & 0.446 & 0.438 & 0.496 & 0.481 & 0.551 & 0.502 & 0.444 & 0.439 & 0.442 & 0.433 \\
            & 720 & 0.468 & 0.471 & 0.540 & 0.513 & 0.464 & 0.471 & 0.614 & 0.551 & 0.653 & 0.554 & \textcolor{red}{\textbf{0.458}} &\textcolor{blue}{\textbf{ 0.467}} & 0.477 & 0.456 \\
            & Avg & 0.433 & \textcolor{blue}{\textbf{0.435}} & 0.508 & 0.479 & 0.432 & 0.436 & 0.486 & 0.452 & 0.504 & 0.473 & \textcolor{red}{\textbf{0.431}} & 0.437 & 0.427 & 0.426 \\
            \hline
            
        \multirow{5}{*}{\begin{turn}{90} 
         ETTh2 
         \end{turn}} 
            & 96  & \textcolor{red}{\textbf{0.284}} & \textcolor{blue}{\textbf{0.338}} & 0.365 & 0.406 & 0.292 & 0.348 & 0.285 & 0.343 & 0.310 & 0.367 & 0.290 & 0.347 & 0.285 & 0.342 \\
\hline
    \end{tabular}
    \caption{OFA Long-term Forecasting Task: We highlight the best performer in MSE index by \textcolor{red}{\textbf{Red}} and MAE by \textcolor{blue}{\textbf{Blue}}}
    \label{tab:ofa_full_forecasting}
\end{table*}

\begin{table*}[ht]
    \centering
    \begin{tabular}{cc|cc|cc|cc|cc|cc|cc|cc}
         \hline
         \multicolumn{2}{c|}{Method} & \multicolumn{2}{c|}{LLM} &  \multicolumn{2}{c|}{Random} & \multicolumn{2}{c|}{LN} & \multicolumn{2}{c|}{Att} & \multicolumn{2}{c|}{Trans} & \multicolumn{2}{c|}{NoLLM} & \multicolumn{2}{c}{Original} \\
         \multicolumn{2}{c|}{Pred Length} & MSE & MAE & MSE & MAE & MSE & MAE & MSE & MAE & MSE & MAE & MSE & MAE & MSE & MAE \\
         \hline
         
        \multirow{5}{*}{\begin{turn}{90} 
         ETTh1 
         \end{turn}} 
            & 96  & 0.400 & \textcolor{blue}{\textbf{0.400}} & 0.375 & 0.406 & 0.382 & 0.403 & 0.378 & 0.408 & \textcolor{red}{\textbf{0.370}} & 0.401 & 0.373 & 0.401 & 0.362 & 0.392 \\
            & 192 & 0.424 & 0.434 & 0.411 & 0.431 & 0.411 & 0.427 & 0.417 & 0.431 & 0.420 & 0.433 & \textcolor{red}{\textbf{0.406}} & \textcolor{blue}{\textbf{0.424}} & 0.398 & 0.418 \\
            & 336 & 0.435 & 0.448 & \textcolor{red}{\textbf{0.431}} & \textcolor{blue}{\textbf{0.442}} & \textcolor{red}{\textbf{0.431}} & 0.446 & 0.446 & 0.457 & 0.446 & 0.457 & 0.432 & 0.445 & 0.430 & 0.427 \\
            & 720 & 0.447 & 0.474 & 0.450 & 0.470 & 0.478 & 0.490 & \textcolor{red}{\textbf{0.432}} & 0.463 & 0.514 & 0.507 & 0.433 & \textcolor{blue}{\textbf{0.456}} & 0.442 & 0.457 \\
            & Avg & 0.426 & 0.439 & 0.417 & 0.437 & 0.426 & 0.441 & 0.424 & 0.440 & 0.438 & 0.449 & \textcolor{red}{\textbf{0.411}} & \textcolor{blue}{\textbf{0.431}} & 0.408 & 0.423 \\
            \hline
            
        \multirow{5}{*}{\begin{turn}{90} 
         ETTh2 
         \end{turn}} 
            & 96  & \textcolor{red}{\textbf{0.275}} & 0.342 & 0.289 & 0.349 & 0.291 & 0.350 & 0.297 & 0.314 & \textcolor{red}{\textbf{0.275}} & \textcolor{blue}{\textbf{0.312}} & 0.292 & 0.349 & 0.268 & 0.328 \\
\hline
    \end{tabular}
    \caption{Time-LLM Long-term Forecasting Task: We highlight the best performer in MSE index by \textcolor{red}{\textbf{Red}} and MAE by \textcolor{blue}{\textbf{Blue}}}
    \label{tab:timellm_forecasting}
\end{table*}

\begin{table*}[ht]
    \centering
    \begin{tabular}{cc|cc|cc|cc|cc|cc|cc|cc}
         \hline
         
         \multicolumn{2}{c|}{Method} & \multicolumn{2}{c|}{LLM} &  \multicolumn{2}{c|}{Random} & \multicolumn{2}{c|}{LN} & \multicolumn{2}{c|}{Att} & \multicolumn{2}{c|}{Trans} & \multicolumn{2}{c|}{NoLLM} & \multicolumn{2}{c}{Original} \\
         \multicolumn{2}{c|}{Pred Length} & MSE & MAE & MSE & MAE & MSE & MAE & MSE & MAE & MSE & MAE & MSE & MAE & MSE & MAE \\
         \hline
         
        \multirow{5}{*}{\begin{turn}{90} 
         ETTh1 
         \end{turn}} 
            & 96   & 0.377 & 0.394 & 0.383 & 0.394 & 0.374 & \textcolor{blue}{\textbf{0.388}} & 0.375 & 0.391 & \textcolor{red}{\textbf{0.371}} & \textcolor{blue}{\textbf{0.388}} & 0.377 & 0.393 & 0.369 & 0.389 \\
            & 192  & 0.428 & 0.425 & 0.428 & 0.422 & 0.424 & \textcolor{blue}{\textbf{0.418}} & \textcolor{red}{\textbf{0.421}} & \textcolor{blue}{\textbf{0.418}} & 0.427 & 0.420 & 0.430 & 0.422 & 0.427 & 0.423 \\
            & 336  & 0.477 & 0.450 & 0.491 & 0.454 & \textcolor{red}{\textbf{0.474}} & \textcolor{blue}{\textbf{0.443}} & 0.479 & 0.455 & 0.475 & \textcolor{blue}{\textbf{0.443}} & \textcolor{red}{\textbf{0.474}} & 0.445 & 0.456 & 0.436 \\
            & 720  & \textcolor{red}{\textbf{0.473}} & \textcolor{blue}{\textbf{0.466}} & 0.484 & 0.474 & 0.478 & \textcolor{blue}{\textbf{0.466}} & 0.485 & 0.471 & 0.483 & 0.469 & 0.478 & 0.469 & 0.479 & 0.467 \\
            & Avg  & 0.439 & 0.434 & 0.446 & 0.436 & \textcolor{red}{\textbf{0.438}} & \textcolor{blue}{\textbf{0.429}} & 0.440 & 0.434 & 0.439 & 0.430 & 0.440 & 0.432 & 0.432 & 0.428 \\
            \hline
            
        \multirow{5}{*}{\begin{turn}{90} 
         ETTh2 
         \end{turn}} 
            & 96   & 0.292 & 0.338 & 0.295 & 0.339 & \textcolor{red}{\textbf{0.289}} & 0.336 & 0.290 & 0.337 & 0.290 & \textcolor{blue}{\textbf{0.335}} & 0.292 & 0.337 & 0.279 & 0.331 \\
\hline
    \end{tabular}
    \caption{CALF Long-term Forecasting Task: We highlight the best performer in MSE index by \textcolor{red}{\textbf{Red}} and MAE by \textcolor{blue}{\textbf{Blue}}}
    \label{tab:my_label}
\end{table*}

\begin{table*}[t]
    \centering
    \begin{tabular}{cc|cc|cc|cc|cc|cc|cc|cc}
         \hline
         
         \multicolumn{2}{c|}{Method} & \multicolumn{2}{c|}{LLM} &  \multicolumn{2}{c|}{Random} & \multicolumn{2}{c|}{LN} & \multicolumn{2}{c|}{Att} & \multicolumn{2}{c|}{Trans} & \multicolumn{2}{c|}{NoLLM} & \multicolumn{2}{c}{Original} \\
         \multicolumn{2}{c|}{Pred Length} & MSE & MAE & MSE & MAE & MSE & MAE & MSE & MAE & MSE & MAE & MSE & MAE & MSE & MAE \\
         \hline
         
        \multirow{5}{*}{\begin{turn}{90} 
         ETTh1 
         \end{turn}} 
            & 96   & 0.380 & 0.407 & 0.386 & 0.414 & \textcolor{red}{\textbf{0.370}} & \textcolor{blue}{\textbf{0.402}} & 0.371 & \textcolor{blue}{\textbf{0.402}} & 0.373 & 0.403 & 0.382 & 0.409 & 0.366 & 0.396 \\
            & 192  & 0.407 & \textcolor{blue}{\textbf{0.425}} & 0.408 & 0.428 & 0.409 & 0.428 & \textcolor{red}{\textbf{0.406}} & \textcolor{blue}{\textbf{0.425}} & 0.412 & 0.429 & 0.409 & 0.430 & 0.401 & 0.420 \\
            & 336  & 0.433 & 0.450 & 0.502 & 0.486 & 0.422 & 0.442 & \textcolor{red}{\textbf{0.420}} & \textcolor{blue}{\textbf{0.438}} & 0.421 & 0.442 & 0.536 & 0.499 & 0.412 & 0.431 \\
            & 720  & 0.467 & 0.483 & 0.585 & 0.530 & \textcolor{red}{\textbf{0.441}} & 0.468 & 0.453 & \textcolor{blue}{\textbf{0.475}} & 0.492 & 0.500 & 0.506 & 0.494 & 0.440 & 0.458 \\
            & Avg  & 0.422 & 0.442 & 0.470 & 0.464 & \textcolor{red}{\textbf{0.410}} & \textcolor{blue}{\textbf{0.435}} & 0.412 & \textcolor{blue}{\textbf{0.435}} & 0.424 & 0.442 & 0.458 & 0.458 & 0.406 & 0.427 \\
            \hline
            
        \multirow{5}{*}{\begin{turn}{90} 
         ETTh2 
         \end{turn}} 
            & 96   & 0.300 & 0.354 & 0.325 & 0.366 & 0.291 & 0.348 & 0.294 & 0.348 & \textcolor{red}{\textbf{0.286}} & \textcolor{blue}{\textbf{0.347}} & 0.310 & 0.361 & 0.278 & 0.340 \\
\hline
    \end{tabular}
    \caption{S2IP Long-term Forecasting Task: We highlight the best performer in MSE index by \textcolor{red}{\textbf{Red}} and MAE by \textcolor{blue}{\textbf{Blue}}}
    \label{tab:my_label}
\end{table*}

\begin{table*}[hbt!]
    \centering
    \vspace*{5cm}
    \begin{tabular}{cc|cc|cc|cc|cc|cc|cc|cc}
         \hline
         \multicolumn{2}{c|}{Method} & \multicolumn{2}{c|}{LLM} & \multicolumn{2}{c|}{Random} & \multicolumn{2}{c|}{LN} & \multicolumn{2}{c|}{Att} & \multicolumn{2}{c|}{Trans} & \multicolumn{2}{c|}{NoLLM} & \multicolumn{2}{c}{Original} \\
         \multicolumn{2}{c|}{Mask Ratio} & MSE & MAE & MSE & MAE & MSE & MAE & MSE & MAE & MSE & MAE & MSE & MAE & MSE & MAE \\
         \hline
         
         \multirow{5}{*}{\begin{turn}{90} 
         ECL 
         \end{turn}} 
            & 12.5\% & 0.073 & 0.189 & \textcolor{red}{\textbf{0.072}} & \textcolor{blue}{\textbf{0.188}} & \textcolor{red}{\textbf{0.072}} & 0.190 & 0.077 & 0.194 & 0.073 & \textcolor{blue}{\textbf{0.188}} & 0.082 & 0.202 & 0.080 & 0.194 \\
            & 25\%   & \textcolor{red}{\textbf{0.081}} & \textcolor{blue}{\textbf{0.199}} & 0.087 & \textcolor{blue}{\textbf{0.199}} & 0.086 & 0.207 & 0.084 & 0.204 & 0.082 & 0.201 & 0.093 & 0.216 & 0.087 & 0.203 \\
            & 37.5\% & \textcolor{red}{\textbf{0.089}} & \textcolor{blue}{\textbf{0.208}} & 0.090 & 0.209 & 0.097 & 0.221 & 0.091 & 0.212 & 0.091 & 0.213 & 0.103 & 0.228 & 0.094 & 0.211 \\
            & 50\%   & \textcolor{red}{\textbf{0.098}} & \textcolor{blue}{\textbf{0.219}} & 0.099 & 0.220 & 0.110 & 0.235 & 0.102 & 0.225 & 0.101 & 0.224 & 0.115 & 0.241 & 0.101 & 0.220 \\
            & Avg    & \textcolor{red}{\textbf{0.085}} & \textcolor{blue}{\textbf{0.204}} & 0.087 & \textcolor{blue}{\textbf{0.204}} & 0.091 & 0.213 & 0.088 & 0.209 & 0.086 & 0.206 & 0.098 & 0.222 & 0.090 & 0.207 \\

            \hline
        \multirow{5}{*}{\begin{turn}{90} 
         ETTh1 
         \end{turn}} 
            & 12.5\% & 0.042 & 0.144 & \textcolor{red}{\textbf{0.041}} & \textcolor{blue}{\textbf{0.143}} & 0.085 & 0.192 & 0.072 & 0.181 & 0.074 & 0.182 & 0.087 & 0.193 & 0.043 &0.140 \\
\hline
    \end{tabular}
    \caption{OFA Imputation Task: We highlight the best performer in MSE index by \textcolor{red}{\textbf{Red}} and MAE by \textcolor{blue}{\textbf{Blue}}}
    \label{tab:OFA_imputation_full_experiments}
\end{table*}

\begin{table*}[hbt!]
    \centering
    \vspace*{5cm}
    \begin{tabular}{cc|cc|cc|cc|cc|cc|cc|cc}
         \hline
         \multicolumn{2}{c|}{Method} & \multicolumn{2}{c|}{LLM} &  \multicolumn{2}{c|}{Random} & \multicolumn{2}{c|}{LN} & \multicolumn{2}{c|}{Att} & \multicolumn{2}{c|}{Trans} & \multicolumn{2}{c|}{NoLLM} & \multicolumn{2}{c}{Original} \\
         \multicolumn{2}{c|}{Mask Ratio} & MSE & MAE & MSE & MAE & MSE & MAE & MSE & MAE & MSE & MAE & MSE & MAE & MSE & MAE\\
         \hline
         
         \multirow{5}{*}{\begin{turn}{90} 
         ECL 
         \end{turn}} 
            & 12.5\% & 0.115 & 0.236 & 0.142 & 0.261 & 0.104 & 0.222 & 0.132 & 0.252 & \textcolor{red}{\textbf{0.082}} & \textcolor{blue}{\textbf{0.205}} & 0.140 & 0.269 & - & - \\
            & 25\%   & 0.128 & 0.247 & 0.193 & 0.306 & \textcolor{red}{\textbf{0.103}} & \textcolor{blue}{\textbf{0.226}} & 0.144 & 0.260 & 0.120 & 0.246 & 0.143 & 0.256 & - & - \\
            & 37.5\% & 0.144 & 0.258 & 0.210 & 0.318 & \textcolor{red}{\textbf{0.110}} & \textcolor{blue}{\textbf{0.231}} & 0.168 & 0.283 & 0.138 & 0.252 & 0.179 & 0.267 & - & - \\
            & 50\%   & 0.167 & 0.280 & 0.252 & 0.350 & \textcolor{red}{\textbf{0.100}} & \textcolor{blue}{\textbf{0.221}} & 0.178 & 0.290 & 0.142 & 0.265 & 0.181 & 0.290 & - & - \\
            & Avg    & 0.138 & 0.255 & 0.199 & 0.309 & \textcolor{red}{\textbf{0.104}} & \textcolor{blue}{\textbf{0.225}} & 0.155 & 0.271 & 0.121 & 0.242 & 0.161 & 0.271 & - & - \\

            \hline
        \multirow{5}{*}{\begin{turn}{90} 
         ETTh1 
         \end{turn}} 
            & 12.5\% & 0.273 & \textcolor{blue}{\textbf{0.257}} & \textcolor{red}{\textbf{0.167}} & 0.280 & 0.220 & 0.280 & 0.321 & 0.331 & 0.223 & 0.284 & 0.325 & 0.333 & - & - \\
\hline
    \end{tabular}
    \caption{CALF Imputation Task: We highlight the best performer in MSE index by \textcolor{red}{\textbf{Red}} and MAE by \textcolor{blue}{\textbf{Blue}}}
    \label{tab:CALF_imputation_full_experiments}
\end{table*}

\begin{table*}[ht]
    \centering
    \vspace*{5cm}
    \begin{tabular}{cc|cc|cc|cc|cc|cc|cc|cc}
         \hline
         \multicolumn{2}{c|}{Method} & \multicolumn{2}{c|}{LLM} &  \multicolumn{2}{c|}{Random} & \multicolumn{2}{c|}{LN} & \multicolumn{2}{c|}{Att} & \multicolumn{2}{c|}{Trans} & \multicolumn{2}{c|}{NoLLM} & \multicolumn{2}{c}{Original} \\
         \multicolumn{2}{c|}{Mask Ratio} & MSE & MAE & MSE & MAE & MSE & MAE & MSE & MAE & MSE & MAE & MSE & MAE & MSE & MAE\\
         \hline
         
         \multirow{5}{*}{\begin{turn}{90} 
         ECL 
         \end{turn}} 
            & 12.5\% & \textcolor{red}{\textbf{0.010}} & 0.084 & \textcolor{red}{\textbf{0.010}} & \textcolor{blue}{\textbf{0.083}} & 0.025 & 0.131 & 0.023 & 0.124 & 0.069 & 0.187 & 0.025 & 0.133 & - & - \\
            & 25\%   & 0.011 & 0.088 & \textcolor{red}{\textbf{0.009}} & \textcolor{blue}{\textbf{0.082}} & 0.024 & 0.127 & 0.020 & 0.116 & 0.018 & 0.112 & 0.022 & 0.121 & - & - \\
            & 37.5\% & \textcolor{red}{\textbf{0.008}} & \textcolor{blue}{\textbf{0.077}} & 0.009 & 0.083 & 0.022 & 0.122 & 0.019 & 0.110 & 0.016 & 0.101 & 0.021 & 0.117 & - & - \\
            & 50\%   & 0.009 & 0.081 & \textcolor{red}{\textbf{0.007}} & \textcolor{blue}{\textbf{0.073}} & 0.020 & 0.112 & 0.019 & 0.113 & 0.025 & 0.133 & 0.018 & 0.108 & - & - \\
            & Avg    & 0.009 & 0.082 & \textcolor{red}{\textbf{0.008}} & \textcolor{blue}{\textbf{0.080}} & 0.022 & 0.123 & 0.020 & 0.116 & 0.032 & 0.133 & 0.021 & 0.119 & - & - \\
            \hline
            
        \multirow{5}{*}{\begin{turn}{90} 
         ETTh1 
         \end{turn}} 
            & 12.5\% & 0.030 & 0.117 & \textcolor{red}{\textbf{0.013}} & \textcolor{blue}{\textbf{0.076}} & 0.083 & 0.195 & 0.069 & 0.176 & 0.015 & 0.084 & 0.011 & 0.071 & - & - \\
\hline
    \end{tabular}
    \caption{Time-LLM Imputation Task: We highlight the best performer in MSE index by \textcolor{red}{\textbf{Red}} and MAE by \textcolor{blue}{\textbf{Blue}}}
    \label{tab:TimeLLM_imputation_full_experiments}
\end{table*}

\begin{table*}[ht]
    \centering
    \vspace*{5cm}
    \begin{tabular}{cc|cc|cc|cc|cc|cc|cc|cc}
         \hline
         \multicolumn{2}{c|}{Method} & \multicolumn{2}{c|}{LLM} &  \multicolumn{2}{c|}{Random} & \multicolumn{2}{c|}{LN} & \multicolumn{2}{c|}{Att} & \multicolumn{2}{c|}{Trans} & \multicolumn{2}{c|}{NoLLM} & \multicolumn{2}{c}{Original}\\
         \multicolumn{2}{c|}{Mask Ratio} & MSE & MAE & MSE & MAE & MSE & MAE & MSE & MAE & MSE & MAE & MSE & MAE & MSE & MAE\\
         \hline
         
         \multirow{5}{*}{\begin{turn}{90} 
         ECL 
         \end{turn}} 
            & 12.5\% & 0.065 & 0.178 & \textcolor{red}{\textbf{0.061}} & \textcolor{blue}{\textbf{0.171}} & 0.070 & 0.188 & 0.065 & 0.181 & 0.066 & 0.183 & 0.064 & 0.175 & - & - \\
            & 25\%   & 0.079 & 0.199 & 0.075 & 0.194 & 0.075 & 0.198 & 0.069 & 0.186 & 0.070 & 0.190 & \textcolor{red}{\textbf{0.066}} & \textcolor{blue}{\textbf{0.179}} & - & - \\
            & 37.5\% & 0.089 & 0.213 & 0.093 & 0.216 & 0.080 & 0.212 & 0.073 & 0.197 & 0.072 & 0.193 & \textcolor{red}{\textbf{0.068}} & \textcolor{blue}{\textbf{0.183}} & - & - \\
            & 50\%   & 0.098 & 0.228 & 0.097 & 0.227 & 0.088 & 0.223 & 0.077 & 0.202 & 0.078 & 0.206 & \textcolor{red}{\textbf{0.071}} & \textcolor{blue}{\textbf{0.191}} & - & - \\
            & Avg    & 0.083 & 0.204 & 0.081 & 0.202 & 0.078 & 0.205 & 0.071 & 0.191 & 0.071 & 0.193 & \textcolor{red}{\textbf{0.067}} & \textcolor{blue}{\textbf{0.182}} & - & - \\
            \hline
            
        \multirow{5}{*}{\begin{turn}{90} 
         ETTh1 
         \end{turn}} 
            & 12.5\% & 0.050 & 0.147 & 0.050 & 0.147 & 0.056 & 0.149 & 0.648 & 0.479 & \textcolor{red}{\textbf{0.018}} & \textcolor{blue}{\textbf{0.092}} & 0.099 & 0.199 & - & - \\
\hline
    \end{tabular}
    \caption{S2IP Imputation Task: We highlight the best performer in MSE index by \textcolor{red}{\textbf{Red}} and MAE by \textcolor{blue}{\textbf{Blue}}}
    \label{tab:S2IP_imputation_full_experiments}
\end{table*}

\begin{table*}[hbt!]
    \centering
    \tabcolsep=0.11cm
    \begin{tabular}{c|ccc|ccc|ccc|ccc|ccc|c}
         \hline
         \multirow{2}{*}{Method} & \multicolumn{3}{c|}{SMD} &  \multicolumn{3}{c|}{MSL} & \multicolumn{3}{c|}{SMAP} & \multicolumn{3}{c|}{SWaT} & \multicolumn{3}{c|}{PSM} & \multirow{2}{*}{Win} \\
          & Pr & Re & F1 & Pr & Re & F1 & Pr & Re & F1 & Pr & Re & F1 & Pr & Re & F1 \\
         \hline
         LLM        & 87.66 & 81.07 & 84.23 & 82.25 & 82.62 & \textcolor{red}{\textbf{82.43}} & 90.10 & 55.75 & 68.88 & 92.12 & 93.04 & 92.58 & 98.26 & 95.92 & 97.08  & 1 \\
         Random     & 87.74 & 81.19 & 84.34 & 81.16 & 81.17 & 81.17 & 90.86 & 56.74 & \textcolor{red}{\textbf{69.22}} & 92.12 & 92.89 & 92.51 & 98.26 & 95.92 & 97.08  & 1 \\
         LN         & 88.79 & 83.98 & \textcolor{red}{\textbf{86.78}} & 79.20 & 75.87 & 77.50 & 89.30 & 52.42 & 66.06 & 92.11 & 92.78 & 92.45 & 98.64 & 95.80 & \textcolor{red}{\textbf{97.20}} & 2 \\
         Att        & 88.47 & 84.26 & 86.32 & 78.25 & 76.80 & 77.52 & 89.71 & 55.93 & 68.91 & 90.92 & 94.85 & \textcolor{red}{\textbf{92.84}} & 98.55 & 91.64 & 94.97 & 1 \\
         Trans      & 88.14 & 83.83 & 85.93 & 77.90 & 76.02 & 76.95 & 89.66 & 53.59 & 67.08 & 88.75 & 91.04 & 89.88 & 99.03 & 94.67 & 96.80 & 0 \\
         NoLLM      & 87.34 & 80.70 & 83.89 & 82.14 & 81.87 & 82.00 & 89.40 & 52.46 & 66.12 & 91.25 & 92.01 & 91.63 & 98.79 & 90.97 & 94.72 & 0 \\
         Original   & 88.89 & 84.98 & 86.89 & 82.00 & 82.91 & 82.45 & 90.60 & 60.95 & 72.88 & 92.20 & 96.34 & 94.23 & 98.62 & 95.68 & 97.13 & - \\
         \hline
    \end{tabular}
    \caption{OFA Anomaly Detection Task}
    \label{tab:my_label}
\end{table*}

\begin{table*}[ht]
    \centering
    \tabcolsep=0.11cm
    \begin{tabular}{c|ccc|ccc|ccc|ccc|ccc|c}
         \hline
         \multirow{2}{*}{Method} & \multicolumn{3}{c|}{SMD} &  \multicolumn{3}{c|}{MSL} & \multicolumn{3}{c|}{SMAP} & \multicolumn{3}{c|}{SWaT} & \multicolumn{3}{c|}{PSM} & \multirow{2}{*}{Win} \\
          & Pr & Re & F1 & Pr & Re & F1 & Pr & Re & F1 & Pr & Re & F1 & Pr & Re & F1  \\
         \hline
         LLM         & 86.74 & 79.83 & 83.14 & 83.13 & 82.95 & \textcolor{red}{\textbf{83.04}} & 89.55 & 54.09 & 67.44 & 95.97 & 92.37 & \textcolor{red}{\textbf{94.13}} & 98.56 & 91.23 & 94.75 & 2 \\
         Random      & 86.88 & 80.39 & 83.50 & 83.21 & 82.87 & 83.02 & 89.56 & 54.03 & 67.40 & 95.96 & 92.37 & \textcolor{red}{\textbf{94.13}} & 98.78 & 94.56 & 96.63 & 1 \\
         LN          & 88.32 & 80.65 & 84.31 & 77.80 & 71.58 & 74.56 & 93.58 & 52.78 & 66.19 & 95.68 & 86.80 & 91.02 & 98.34 & 91.09 & 94.58 & 0 \\
         Att         & 86.83 & 80.75 & 83.68 & 83.13 & 83.01 & \textcolor{red}{\textbf{83.04}} & 89.73 & 55.93 & \textcolor{red}{\textbf{68.87}} & 95.31 & 80.05 & 87.02 & 98.81 & 94.75 & \textcolor{red}{\textbf{96.76}} & 3 \\
         Trans       & 88.07 & 81.93 & \textcolor{red}{\textbf{84.41}} & 77.01 & 68.74 & 72.64 & 90.77 & 52.86 & 66.27 & 95.33 & 80.39 & 87.24 & 98.25 & 92.08 & 95.07 & 1 \\
         NoLLM       & 86.58 & 79.37 & 82.82 & 83.19 & 82.90 & \textcolor{red}{\textbf{83.04}} & 89.49 & 53.98 & 67.32 & 95.95 & 92.38 & \textcolor{red}{\textbf{94.13}} & 98.84 & 94.64 & 96.70 & 2 \\
         Original    & - & - & - & - & - & - & - & - & - & - & - & - & - & - & - & - \\
         \hline
    \end{tabular}
    \caption{CALF Anomaly Detection Task}
    \label{tab:my_label}
\end{table*}

\begin{table*}[ht]
    \centering
     \tabcolsep=0.11cm
    \begin{tabular}{c|ccc|ccc|ccc|ccc|ccc|c}
         \hline
         \multirow{2}{*}{Method} & \multicolumn{3}{c|}{SMD} &  \multicolumn{3}{c|}{MSL} & \multicolumn{3}{c|}{SMAP} & \multicolumn{3}{c|}{SWaT} & \multicolumn{3}{c|}{PSM} & \multirow{2}{*}{Win} \\
         & Pr & Re & F1 & Pr & Re & F1 & Pr & Re & F1 & Pr & Re & F1 & Pr & Re & F1  \\
         \hline
         LLM        & 86.45 & 77.84 & 81.17 & 80.74 & 75.31 & 77.93 & 90.77 & 55.57 & 68.94 & 99.90 & 68.74 & 81.45 & 98.59 & 91.98 & 95.17 & 0 \\
         Random     & 86.69 & 78.03 & 81.44 & 80.66 & 75.31 & 77.98 & 90.32 & 54.48 & 68.10 & 99.96 & 68.73 & \textcolor{red}{\textbf{81.46}} & 98.77 & 92.03 & 95.44 & 1 \\
         LN         & 87.81 & 78.67 & \textcolor{red}{\textbf{82.99}} & 79.81 & 76.49 & 78.11 & 91.46 & 57.68 & \textcolor{red}{\textbf{70.78}} & 99.97 & 65.62 & 79.23 & 98.30 & 94.67 & 96.45 & 2 \\
         Att        & 87.25 & 74.49 & 80.35 & 76.24 & 62.32 & 68.52 & 89.13 & 54.61 & 67.80 & 99.96 & 66.79 & 80.08 & 98.54 & 94.66 & \textcolor{red}{\textbf{96.85}} & 1 \\
         Trans      & 87.10 & 75.10 & 80.66 & 76.90 & 62.64 & 69.04 & 88.81 & 53.44 & 66.31 & 99.92 & 67.42 & 80.94 & 98.35 & 94.32 & 96.29 & 0 \\
         NoLLM      & 86.98 & 78.18 & 81.28 & 81.35 & 76.84 & \textcolor{red}{\textbf{79.45}} & 90.61 & 55.13 & 68.83 & 99.98 & 65.62 & 79.24 & 98.13 & 94.22 & 96.14 & 1 \\
         Original   & - & - & - & - & - & - & - & - & - & - & - & - & - & - & - & - \\
         \hline
    \end{tabular}
    \caption{S2IP Anomaly Detection Task}
    \label{tab:my_label}
\end{table*}

\begin{table*}[ht]
    \centering
     \tabcolsep=0.11cm
    \begin{tabular}{c|ccc|ccc|ccc|ccc|ccc|c}
         \hline
         \multirow{2}{*}{Method} & \multicolumn{3}{c|}{SMD} &  \multicolumn{3}{c|}{MSL} & \multicolumn{3}{c|}{SMAP} & \multicolumn{3}{c|}{SWaT} & \multicolumn{3}{c|}{PSM} & \multirow{2}{*}{Win} \\
          & Pr & Re & F1 & Pr & Re & F1 & Pr & Re & F1 & Pr & Re & F1 & Pr & Re & F1 \\
         \hline
         GPT2       & 82.10 & 82.16 & 82.13 & 51.08 & 96.23 & \textcolor{red}{\textbf{66.74}} & 90.42 & 52.66 & 66.55 & 88.57 & 84.07 & 86.26 & 99.52 & 78.86 & 88.01 & 1 \\
         Random     & 81.81 & 82.63 & 82.22 & 50.72 & 96.07 & 66.29 & 91.81 & 52.59 & \textcolor{red}{\textbf{66.88}} & 89.66 & 84.88 & \textcolor{red}{\textbf{87.11}} & 98.82 & 75.63 & 85.68 & 2 \\
         LN         & 82.08 & 83.23 & \textcolor{red}{\textbf{82.65}} & 50.67 & 96.23 & 66.40 & 93.24 & 50.63 & 65.62 & 85.68 & 83.42 & 84.54 & 99.56 & 79.30 & 88.28 & 1 \\
         Att        & 87.47 & 74.57 & 80.50 & 40.31 & 96.27 & 56.70 & 93.24 & 50.62 & 65.62 & 88.74 & 77.21 & 82.57 & 98.72 & 74.60 & 84.96 & 0 \\
         Trans      & 81.32 & 81.05 & 81.18 & 49.53 & 96.23 & 65.40 & 93.15 & 50.62 & 65.60 & 85.63 & 83.51 & 84.56 & 99.51 & 79.77 & \textcolor{red}{\textbf{88.56}} & 1 \\
         NoLLM      & 81.76 & 82.33 & 82.05 & 47.20 & 96.28 & 63.33 & 86.67 & 53.20 & 65.92 & 88.75 & 77.63 & 82.82 & 99.08 & 76.01 & 86.03 & 0 \\
         Original   & - & - & - & - & - & - & - & - & - & - & - & - & - & - & - & - \\
         \hline
    \end{tabular}
    \caption{Time-LLM Anomaly Detection Task}
    \label{tab:my_label}
\end{table*}

\begin{table*}[ht]
    \centering
    \begin{tabular}{c|c|c|c|c|c|c|c}
         \hline
         \multicolumn{8}{c}{One-Fits-All} \\ \hline
         Dataset & LLM &  Random & LN & Att & Trans & NoLLM & Original \\
         \hline
         EthanolConcentration & \textcolor{red}{\textbf{34.2}} & 33.4 & 32.3 & 30.4 & 33.4 & 30.0 & 34.2 \\
         FaceDetection &  67.1 & 67.6 & 67.9 & 66.9 & \textcolor{red}{\textbf{68.6}} & 67.5 & 69.2 \\
         Handwriting & 29.7 & 31.6 & 29.2 & 33.0 & 30.9 & \textcolor{red}{\textbf{38.1}} & 32.7 \\
         Heartbeat & 75.1 &  76.5 & 73.6 & 77.0 & 76.1 & \textcolor{red}{\textbf{77.5}} & 77.2 \\
         JapaneseVowels & 98.1 &  98.1 & \textcolor{red}{\textbf{98.3}} & 97.5 & 98.1 & 97.5 & 98.6 \\
         PEMS-SF & 80.9 & 81.5 & 80.3 & \textcolor{red}{\textbf{86.7}} & 79.7 & 79.7 & 87.9 \\
         SelfRegulationSCP1 & 91.2 &  91.8 & 89.7 & 91.1 & \textcolor{red}{\textbf{92.1}} & 90.4 & 93.2 \\
         SelfRegulationSCP2 & 56.1 & \textcolor{red}{\textbf{57.2}} & 56.1 & 55.1 & 55.3 & 54.4  & 59.4 \\
         SpokenArabicDigits & 99.0 &  98.8 & 97.9 & 98.5 & 98.5 & \textcolor{red}{\textbf{99.1}} & 99.2 \\
         UWaveGestureLibrary & 86.5 &  85.3 & 85.9 & 85.5 & 84.6 & \textcolor{red}{\textbf{87.1}} & 88.1 \\
         \hline
         Average & 71.8 & 72.2 & 71.0 &  \textcolor{red}{\textbf{72.2}} & 71.7 & 72.1 & 74.0 \\
         \hline
         Wins & 1 & 1 & 1 & 1 & 1 & 4 & - \\
         \hline
    \end{tabular}
    \caption{OFA Multivariate Classification}
    \label{tab:multi_classification}
\end{table*}

\begin{table*}[ht]
    \centering
    \begin{tabular}{c|c|c|c|c|c|c|c}
         \hline
         \multicolumn{8}{c}{CALF} \\ \hline
         Dataset & LLM &  Random & LN & Att & Trans & NoLLM & Original \\
         \hline
         EthanolConcentration & 25.8 & 25.8 & \textcolor{red}{\textbf{28.1}} & 27.3 & 27.3 & 26.6 & -  \\
         FaceDetection & 68.1 & 67.3 & 67.7 & 67.7 & 67.7 & \textcolor{red}{\textbf{68.4}} & - \\
         Handwriting & 26.8 & 23.6 & 25.8 & 25.2 & 25.2 & \textcolor{red}{\textbf{26.8}} &  - \\
         Heartbeat & 75.6 & \textcolor{red}{\textbf{75.6}} & 73.1 & 75.6 & 75.6 & 70.7 &  - \\
         JapaneseVowels & 95.1 & 95.4 & \textcolor{red}{\textbf{95.9}} & 95.4 & 95.4 & 95.4 &  - \\
         PEMS-SF & 78.2 & \textcolor{red}{\textbf{78.6}} & 78.6 & 78.0 & 78.0 & 77.4 &  - \\
         SelfRegulationSCP1 & 89.4 & 89.4 & 89.7 & \textcolor{red}{\textbf{90.7}} & \textcolor{red}{\textbf{90.7}} & 89.0 & -  \\
         SelfRegulationSCP2 & 53.8 & 53.9 & 59.4 & \textcolor{red}{\textbf{60.8}} & 59.4 & 51.1 &  - \\
         SpokenArabicDigits & 98.1 & 98.1 & 98.5 & 98.5 & 98.5 & \textcolor{red}{\textbf{98.8}} &  - \\
         UWaveGestureLibrary & 86.5 & 86.5 & 84.6 & 84.6 & 84.6 & \textcolor{red}{\textbf{87.1}} & -  \\
         \hline
         Average & 69.78 & 69.46 & 70.19 &  \textcolor{red}{\textbf{70.43}} & 70.29 & 69.17 & - \\
         \hline
         Wins & 0 & 2 & 2 & 2 & 1 & 4 & - \\
         \hline
    \end{tabular}
    \caption{CALF Multivariate Classification}
    \label{tab:my_label}
\end{table*}

\begin{table*}[ht]
    \centering
    \begin{tabular}{c|c|c|c|c|c|c|c}
         \hline
         \multicolumn{8}{c}{Time-LLM} \\ \hline
         Dataset & LLM &  Random & LN & Att & Trans & NoLLM & Original \\
         \hline
         EthanolConcentration & \textcolor{red}{\textbf{30.0}} & 25.0 & 28.1 & 25.0 & 24.7 & 25.0 & - \\
         FaceDetection & 50.2 & 50.0 & 51.1 & \textcolor{red}{\textbf{51.5}} & 50.2 & 50.4 &  -\\
         Handwriting & \textcolor{red}{\textbf{6.7}} & 5.5 & 5.1 & 3.5 & 3.7 & 4.2  & - \\
         Heartbeat & 72.1 & 72.1 & 72.1 & 72.1 & 72.1 & 72.1  & - \\
         JapaneseVowels & 23.9 & 16.4 & 25.4 & 30.5 & 23.2 & \textcolor{red}{\textbf{42.1}}  &  - \\
         PEMS-SF & 23.1 & 30.6 & 35.8 & 40.8 & 39.3 & \textcolor{red}{\textbf{48.5}}  & - \\
         SelfRegulationSCP1 & \textcolor{red}{\textbf{63.8}} & 49.8 & 50.1 & 50.1 & 50.1 & 62.7  & - \\
         SelfRegulationSCP2 & 50.0 & 50.0 & \textcolor{red}{\textbf{53.8}} & 50.0 & 50.0 & 50.0  & - \\
         SpokenArabicDigits & 22.7 & \textcolor{red}{\textbf{35.2}} & 21.9 & 14.1 & 15.4 & 21.5 & - \\
         UWaveGestureLibrary & 33.4 & \textcolor{red}{\textbf{38.7}} & 29.3 & 14.3 & 37.1 & 31.2  & - \\
         \hline
         Average & 37.6 &  37.3 & 37.3 & 35.2 & 36.6 & \textcolor{red}{\textbf{40.8}} & - \\
         \hline
         Wins & 3 & 2 & 1 & 1 & 0 & 2  & - \\
        \hline
    \end{tabular}
    \caption{Time-LLM Multivariate Classification}
    \label{tab:my_label}
\end{table*}

\begin{table*}[ht]
    \centering
    \begin{tabular}{c|c|c|c|c|c|c|c}
         \hline
        \multicolumn{8}{c}{S2IP} \\ \hline
         Dataset & LLM &  Random & LN & Att & Trans & NoLLM & Original \\
         \hline
         EthanolConcentration & 28.6 & 27.2 & 28.5 & 29.6 & \textcolor{red}{\textbf{30.4}} & 26.9 & - \\
         FaceDetection & 66.5  & 66.4 & 64.1 & \textcolor{red}{\textbf{67.3}} & 65.5 & 62.7  & - \\
         Handwriting & 25.8 & 27.3 & 21.0 & 27.5 & 21.8 & \textcolor{red}{\textbf{31.2}}  & - \\
         Heartbeat & \textcolor{red}{\textbf{72.6}} & 65.9 & 69.2 & 72.2 & 69.2 & 72.2 & - \\
         JapaneseVowels & 93.8 & 92.5 & 94.2 & 92.2 & 92.8 & \textcolor{red}{\textbf{94.3}} & - \\
         PEMS-SF & 80.41 & 81.5 & \textcolor{red}{\textbf{82.9}} & 79.0 & 79.6 & 81.4 \\
         SelfRegulationSCP1 & 85.3 & 87.4 & 84.6 & \textcolor{red}{\textbf{87.7}} & 82.6 & 85.3 & - \\
         SelfRegulationSCP2 & \textcolor{red}{\textbf{57.2}} & 53.3 & 55.0 & 53.8 & 55.0 & 52.7 & - \\
         SpokenArabicDigits & 96.6 & 97.5 & 95.1 & 97.3 & 97.4 & \textcolor{red}{\textbf{97.9}} & - \\
         UWaveGestureLibrary & 82.1 & 84.1 & 82.8 & 80.3 & 82.8 & \textcolor{red}{\textbf{85.9}}  & - \\
         \hline
         Average & 68.9 & 68.7 & 67.6 & 69.0 & 67.7 & \textcolor{red}{\textbf{69.1}} & - \\
         \hline
         Wins & 2 & 0 & 1 & 2 & 1 & 4  & - \\
         \hline
    \end{tabular}
    \caption{S2IP Multivariate Classification}
    \label{tab:my_label}
\end{table*}

\section{Additional Visualization} \label{sec:Additional_Visualization}

\begin{figure*}[hbt!]
    \centering
    
    \includegraphics[width=1\linewidth]{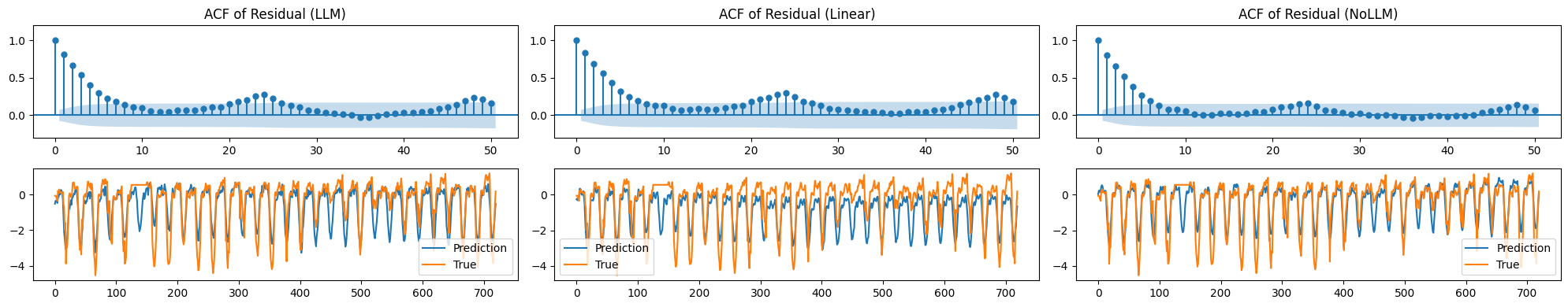}
    \caption{TimeLLM on ETTh1 Dataset: Residual ACF and Forecasting Plots. The Durbin-Watson Statistic for LLM, Linear and NoLLM are 0.3485, 3348, 0.3695}
    \label{fig:TimeLLM_ETTh1_ACF}
\end{figure*}

\begin{figure*}[ht]
    \centering
    \includegraphics[width=1\linewidth]{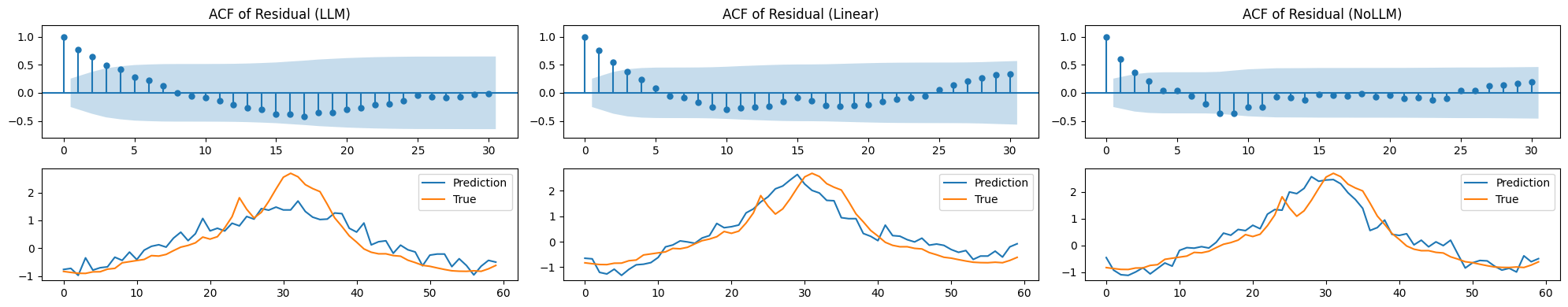}
    \caption{TimeLLM on Illness Dataset: Residual ACF and Forecasting Plots. The Durbin-Watson Statistic for LLM, Linear and NoLLM are 0.1319, 0.1890, 0.2018}
    \label{fig:TimeLLM_Illness_ACF}
\end{figure*}

\begin{figure*}[ht]
    \centering
    \includegraphics[width=1\linewidth]{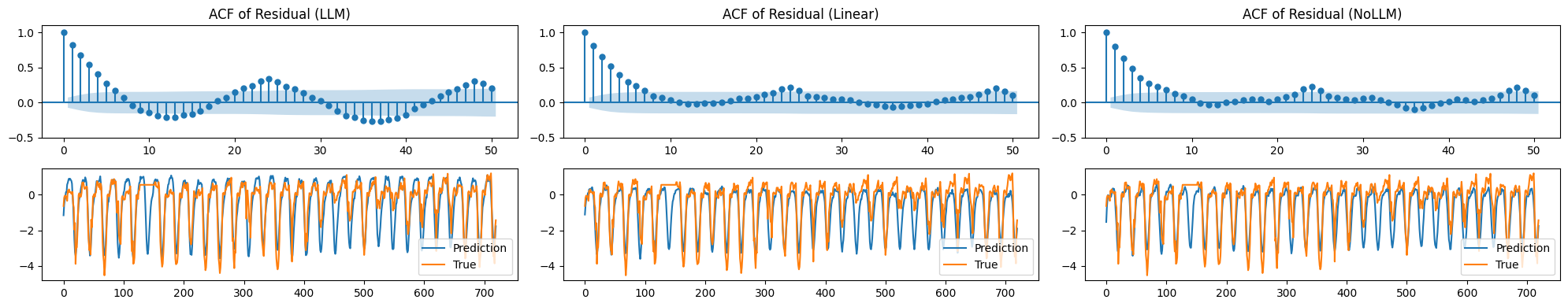}
    \caption{CALF on ETTh1 Dataset: Residual ACF and Forecasting Plots. The Average Durbin-Watson Statistic for LLM, Linear and NoLLM are 0.3450, 0.3588, 0.3641}
    \label{fig:CALF_ETTh1_ACF}
\end{figure*}

\begin{figure*}[ht]
    \centering
    \includegraphics[width=1\linewidth]{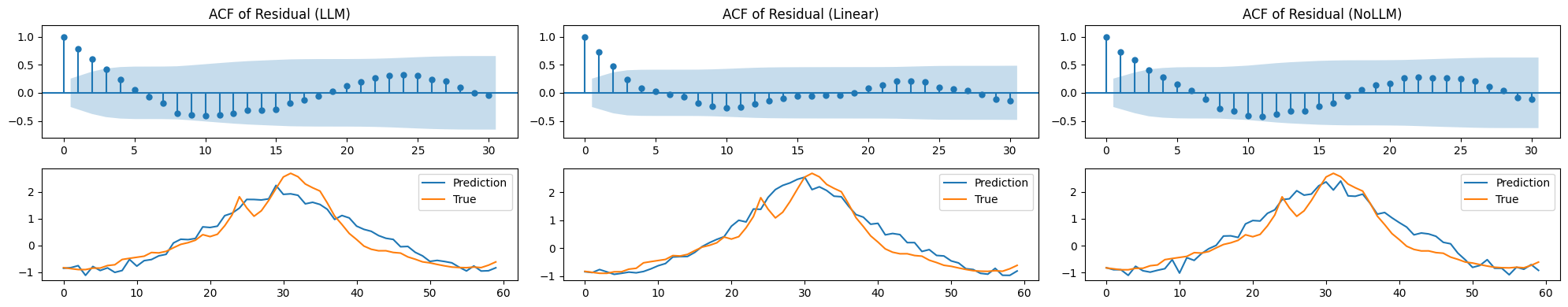}
    \caption{CALF on Illness Dataset: Residual ACF and Forecasting Plots. The Average Durbin-Watson Statistic for LLM, Linear and NoLLM are 0.1464, 0.1542, 0.1825}
    \label{fig:CALF_Illness_ACF}
\end{figure*}

\begin{figure*}[ht]
    \centering
    \includegraphics[width=1\linewidth]{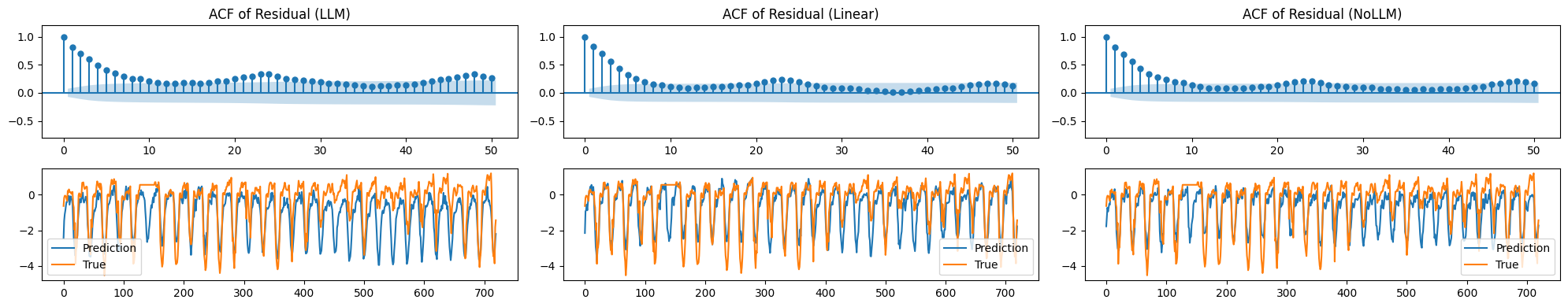}
    \caption{S2IP on ETTh1 Dataset: Residual ACF and Forecasting Plots. The Average Durbin-Watson Statistic for LLM, Linear and NoLLM are 0.3236, 0.3468, 0.3847}
    \label{fig:S2IP_ETTh1_ACF}
\end{figure*}

\begin{figure*}[ht]
    \centering
    \includegraphics[width=1\linewidth]{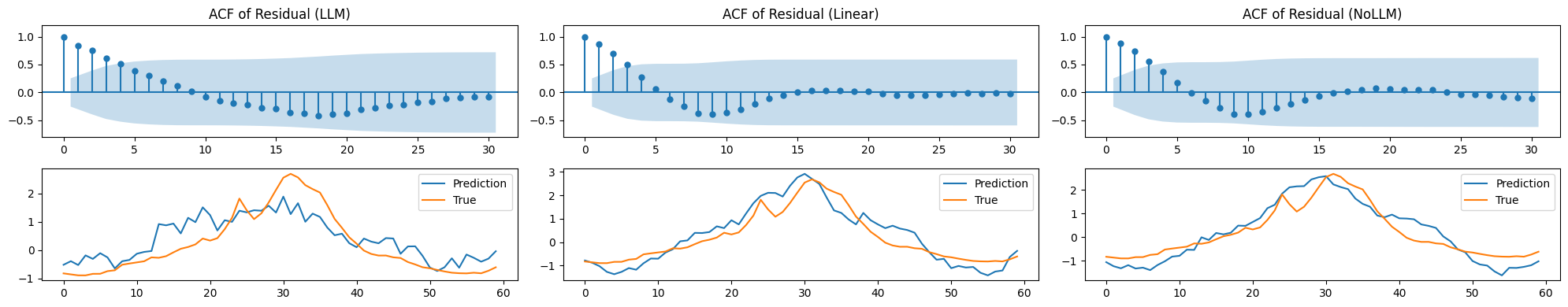}
    \caption{S2IP Illness dataset: Residual ACF and Forecasting Plots. The Average Durbin-Watson Statistic for LLM, Linear and NoLLM are 0.1555, 0.2539, 0.1216}
    \label{fig:S2IP_Illness_ACF}
\end{figure*}

\begin{figure*}[ht]
    \centering
    \includegraphics[width=1\linewidth]{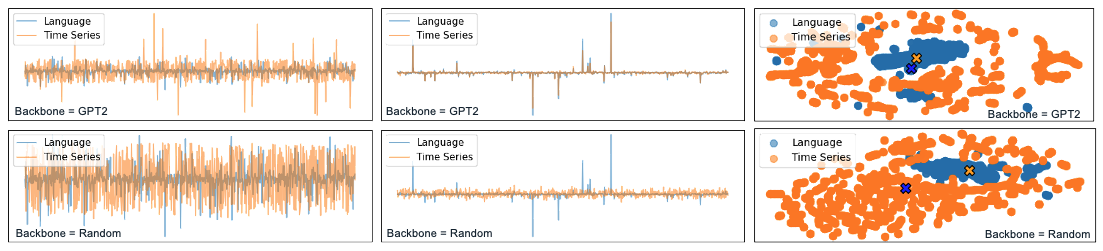}
    \caption{OFA on Illness Dataset: (Column 1): Feature Map before Backbone. (Column 2): Feature Map after Backbone. (Column 3): UMAP Plot of Random and LLM Backbone.}
    \label{fig:OFA_Illness}
\end{figure*}

\begin{figure*}[ht]
    \centering
    \includegraphics[width=1\linewidth]{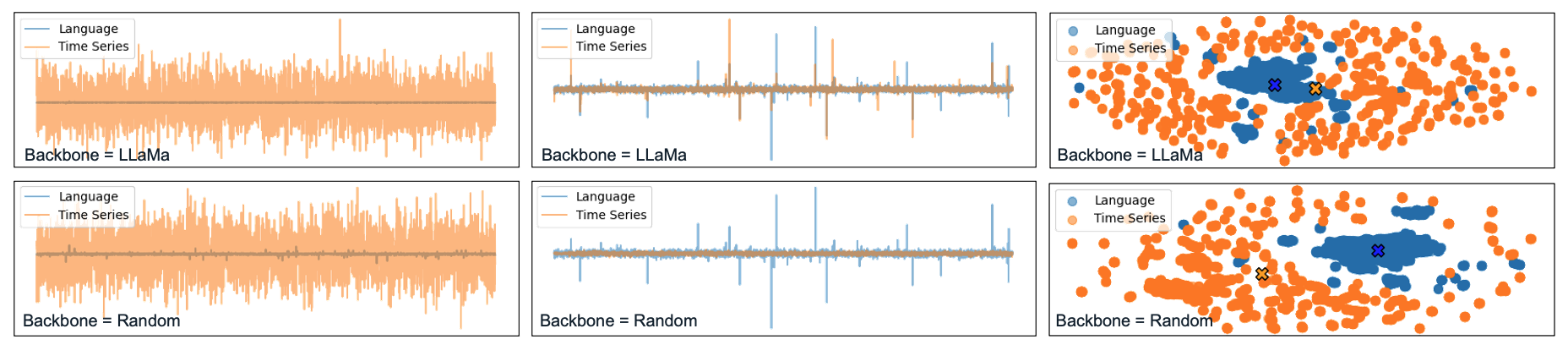}
    \caption{TimeLLM ETTh1 Dataset: (Column 1): Feature Map before Backbone. (Column 2): Feature Map after Backbone. (Column 3): UMAP Plot of Random and LLM Backbone.}
    \label{fig:Time_ETTh1}
\end{figure*}

\begin{figure*}[ht]
    \centering
    \includegraphics[width=1\linewidth]{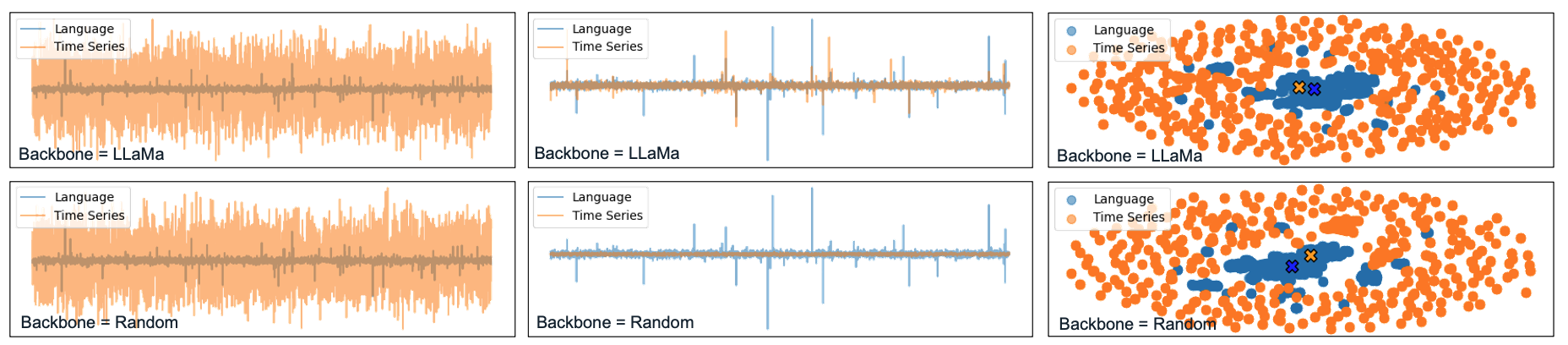}
    \caption{TimeLLM on Illness Dataset: (Column 1): Feature Map before Backbone. (Column 2): Feature Map after Backbone. (Column 3): UMAP Plot of Random and LLM Backbone.}
    \label{fig:Time_Illness}
\end{figure*}

\begin{figure*}[ht]
    \centering
    \includegraphics[width=1\linewidth]{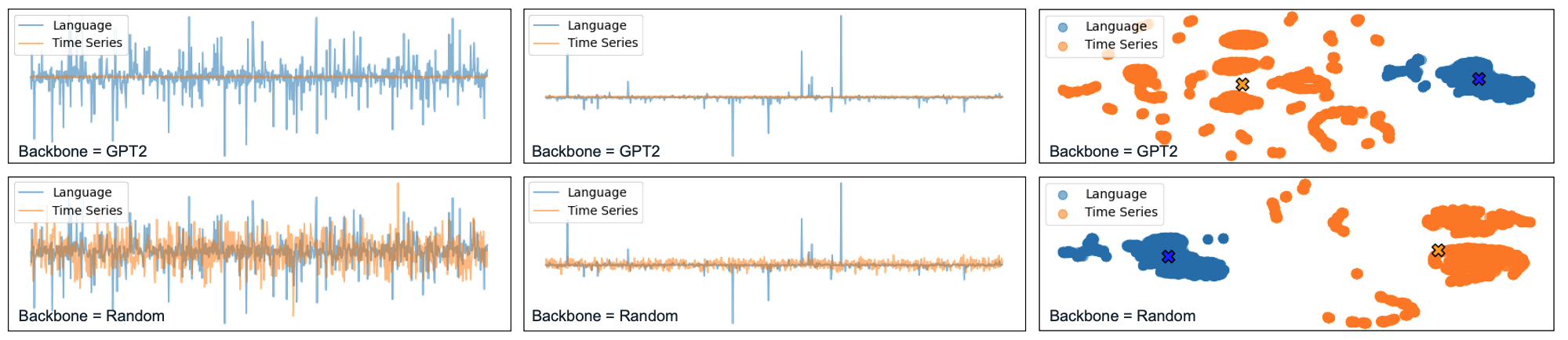}
    \caption{CALF on ETTh1 Dataset: (Column 1): Feature Map before Backbone. (Column 2): Feature Map after Backbone. (Column 3): UMAP Plot of Random and LLM Backbone.}
    \label{fig:CALF_ETTh1}
\end{figure*}

\begin{figure*}[ht]
    \centering
    \includegraphics[width=1\linewidth]{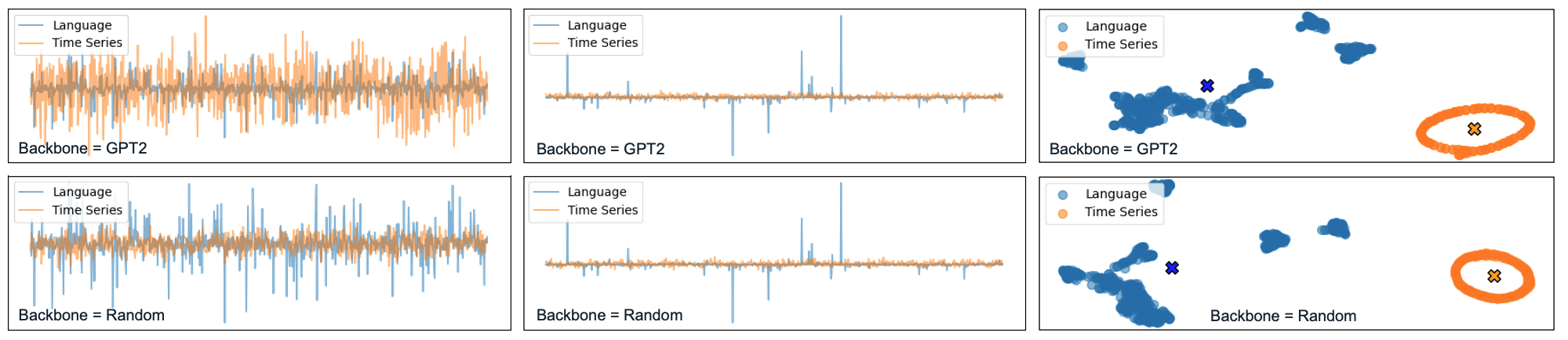}
    \caption{CALF on Illness Dataset: (Column 1): Feature Map before Backbone. (Column 2): Feature Map after Backbone. (Column 3): UMAP Plot of Random and LLM Backbone.}
    \label{fig:CALF_Illness}
\end{figure*}

\begin{figure*}[ht]
    \centering
    \includegraphics[width=1\linewidth]{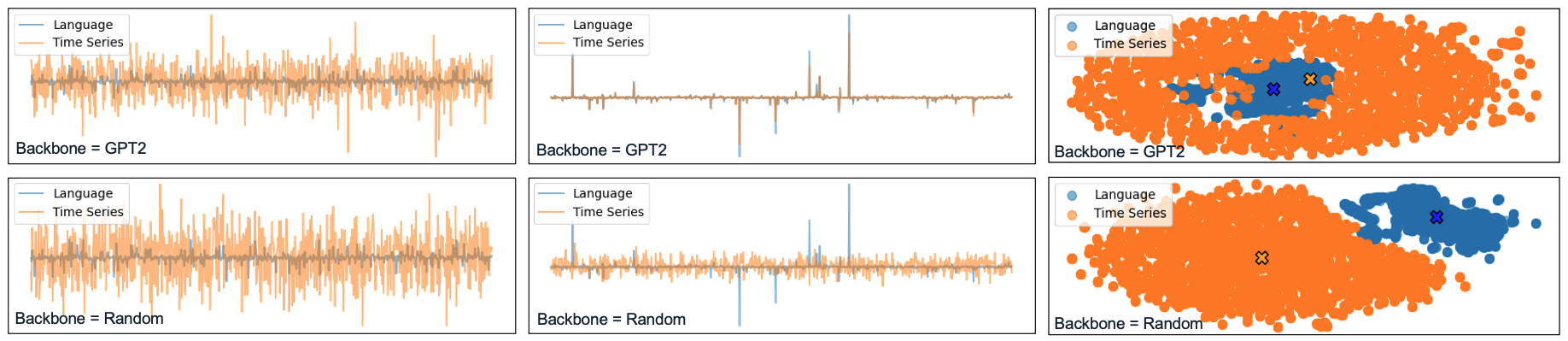}
    \caption{S2IP on ETTh1 Dataset: (Column 1): Feature Map before Backbone. (Column 2): Feature Map after Backbone. (Column 3): UMAP Plot of Random and LLM Backbone.}
    \label{fig:S2IP_ETTh1}
\end{figure*}

\begin{figure*}[ht]
    \centering
    \includegraphics[width=1\linewidth]{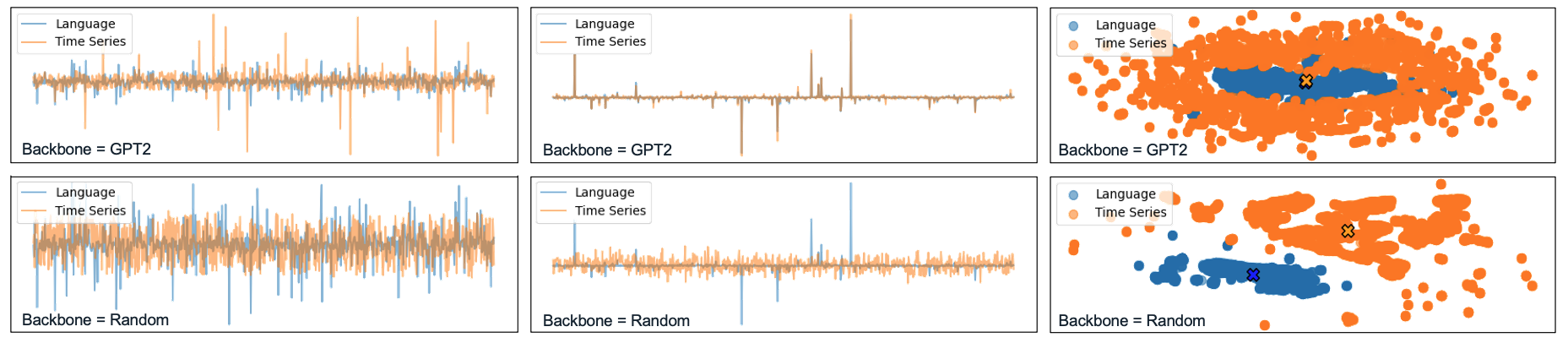}
    \caption{S2IP on Illness Dataset: (Column 1): Feature Map before Backbone. (Column 2): Feature Map after Backbone. (Column 3): UMAP Plot of Random and LLM Backbone.}
    \label{fig:S2IP_Illness}
\end{figure*}


\begin{figure*}[ht]
    \centering
    \begin{subfigure}[t]{0.48\linewidth}
        \centering
        \includegraphics[width=\linewidth]{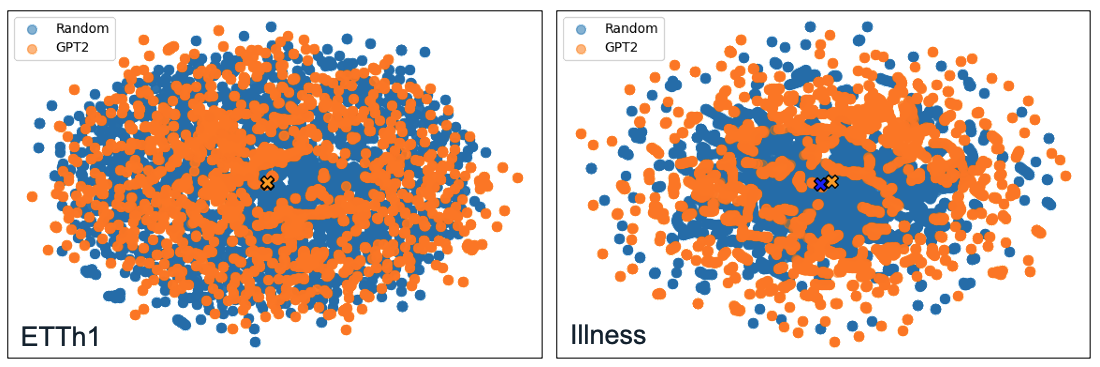}
        \caption{OFA and Random UMAP Representation}
        \label{fig:OFA_Random_UMAP}
    \end{subfigure}
    \begin{subfigure}[t]{0.48\linewidth}
        \centering
        \includegraphics[width=\linewidth]{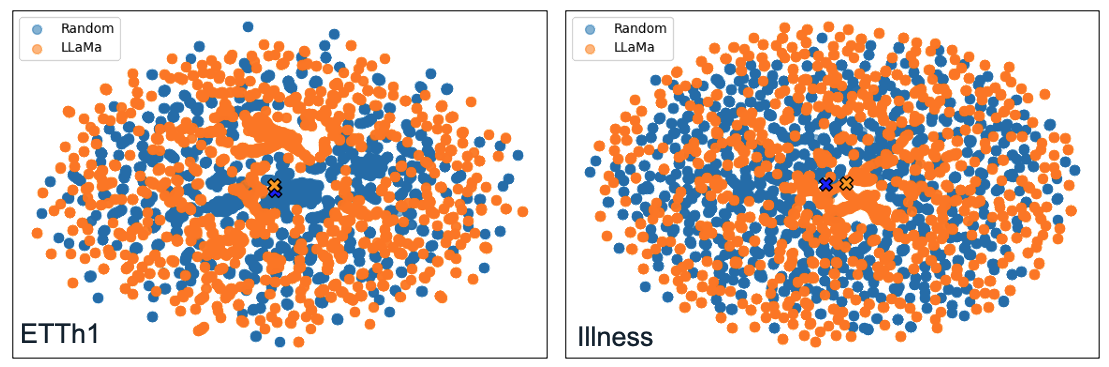}
        \caption{TimeLLM and Random UMAP Representation}
        \label{fig:TimeLLM_Random_UMAP}
    \end{subfigure}
    \vspace{0.5cm}
    
    \begin{subfigure}[t]{0.48\linewidth}
        \centering
        \includegraphics[width=\linewidth]{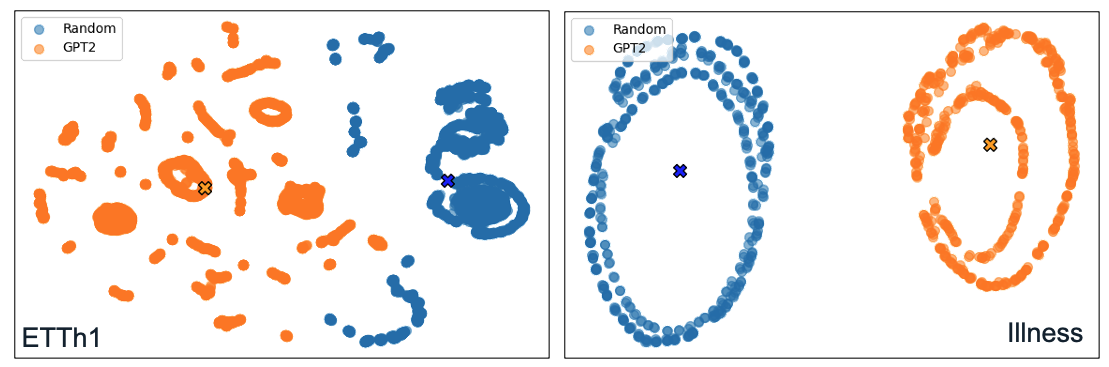}
        \caption{CALF and Random UMAP Representation}
        \label{fig:CALF_Random_UMAP}
    \end{subfigure}
    \begin{subfigure}[t]{0.48\linewidth}
        \centering
        \includegraphics[width=\linewidth]{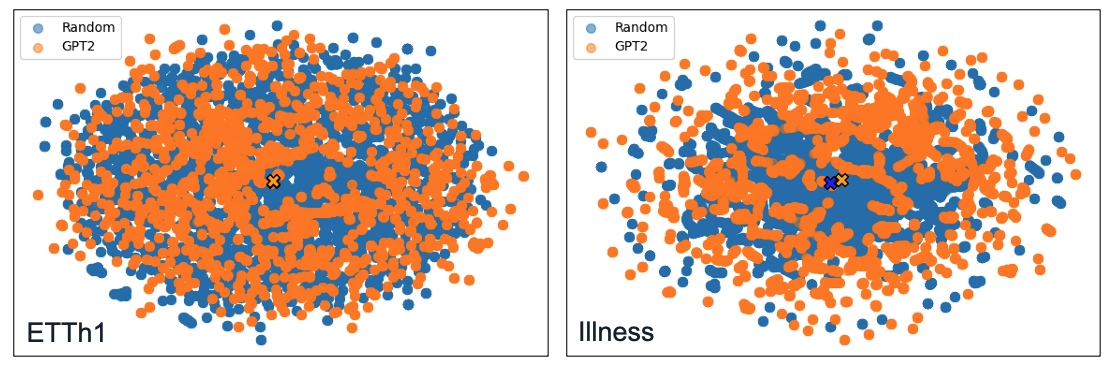}
        \caption{S2IP and Random UMAP Representation}
        \label{fig:S2IP_Random_UMAP}
    \end{subfigure}

    \caption{Time Series Tokens UMAP Plots Comparison between Random Initialization and Pre-trained LLM}
    \label{fig:LLM_Random_UMAP_Comparison}
\end{figure*}

\end{document}